\documentclass[10pt,journal,compsoc]{./IEEEtran}

\ifCLASSOPTIONcompsoc
  \usepackage[nocompress]{cite}
\else
  \usepackage{cite}
\fi
\ifCLASSINFOpdf
  \usepackage[pdftex]{graphicx}
  \graphicspath{{./pics/}}
  \DeclareGraphicsExtensions{.pdf,.jpeg,.png}
\else
\fi
\usepackage{amsmath}
\usepackage{algorithmic}
\usepackage[linesnumbered,ruled]{algorithm2e} 

\usepackage{subcaption}
\usepackage{url}

\usepackage{times}
\usepackage{epsfig}
\usepackage{amssymb}
\usepackage[pagebackref=true,breaklinks=true,letterpaper=true,colorlinks,bookmarks=false]{hyperref}

\usepackage{xcolor}
\usepackage{xspace}
\usepackage{multirow}
\usepackage{booktabs}
\usepackage{placeins}
\usepackage{amssymb}
\usepackage{pifont}

\newcommand{\cmark}{\ding{51}}%
\newcommand{\xmark}{\ding{55}}%

\usepackage{dblfloatfix}    
\usepackage{kantlipsum}

\begin{document}
%
\title{Scale Normalized Image Pyramids with AutoFocus for Object Detection}
%
%
%
%

\author{Bharat Singh, Mahyar Najibi, Abhishek Sharma and Larry S. Davis
\IEEEcompsocitemizethanks{\IEEEcompsocthanksitem B. Singh, M Najibi and L.S. Davis were with the Computer Science Department, University of Maryland, College Park at the time of this research. Email: bharat/najibi/lsd(@cs.umd.edu). Abhishek Sharma was with Gobasco AI Labs at the time of this research. Email:abhisharayiya@gmail.com}
}
\IEEEtitleabstractindextext{%
\begin{abstract}
We present an efficient foveal framework to perform object detection. A scale normalized image pyramid (SNIP) is generated that, like human vision, only attends to objects within a fixed size range at different scales. Such a restriction of objects' size during training affords better learning of object-sensitive filters, and therefore, results in better accuracy. However, the use of an image pyramid increases the computational cost. Hence, we propose an efficient \emph{spatial sub-sampling} scheme which only operates on fixed-size sub-regions likely to contain objects (as object locations are known during training). The resulting approach, referred to as \emph{Scale Normalized Image Pyramid with Efficient Resampling} or SNIPER, yields up to 3$\times$ speed-up during training. Unfortunately, as object locations are unknown during inference, the entire image pyramid still needs processing. To this end, we adopt a coarse-to-fine approach, and predict the locations and extent of object-like regions which will be processed in successive scales of the image pyramid. Intuitively, it's akin to our active human-vision that first \emph{skims over} the field-of-view to spot interesting regions for further processing and only recognizes objects at the right resolution. The resulting algorithm is referred to as \emph{AutoFocus} and results in a 2.5-5$\times$ speed-up during inference when used with SNIP. Code: \url{https://github.com/mahyarnajibi/SNIPER}
\end{abstract}

}

\maketitle

\IEEEdisplaynontitleabstractindextext

%
\IEEEpeerreviewmaketitle
\IEEEraisesectionheading{\section{Introduction}\label{sec:introduction}}

%
%
%
%

\IEEEPARstart{O}{bject}-detection is one of the most popular and widely researched problems in the computer vision community, owing to its application to a myriad of industrial systems, such as autonomous driving, robotics, surveillance, activity-detection, scene-understanding and/or large-scale multimedia analysis. Just like other computer-vision problems, object-detection too has witnessed a quantum leap in its performance with the advent of the deep-learning framework. However, current object-detection systems are far from perfect, or even remotely comparable to humans. This is due to the presence of huge variation in the appearance of objects in terms of lighting, occlusion, viewpoint, deformations, scale, and, to some extent, sensor differences. In this work, we focus on the challenges arising due to \emph{scale variation} and propose solutions to achieve state-of-the-art object-detection performance with \emph{near} real-time latency.

\begin{figure}[t]
    \center
    \includegraphics[width=0.99\linewidth]{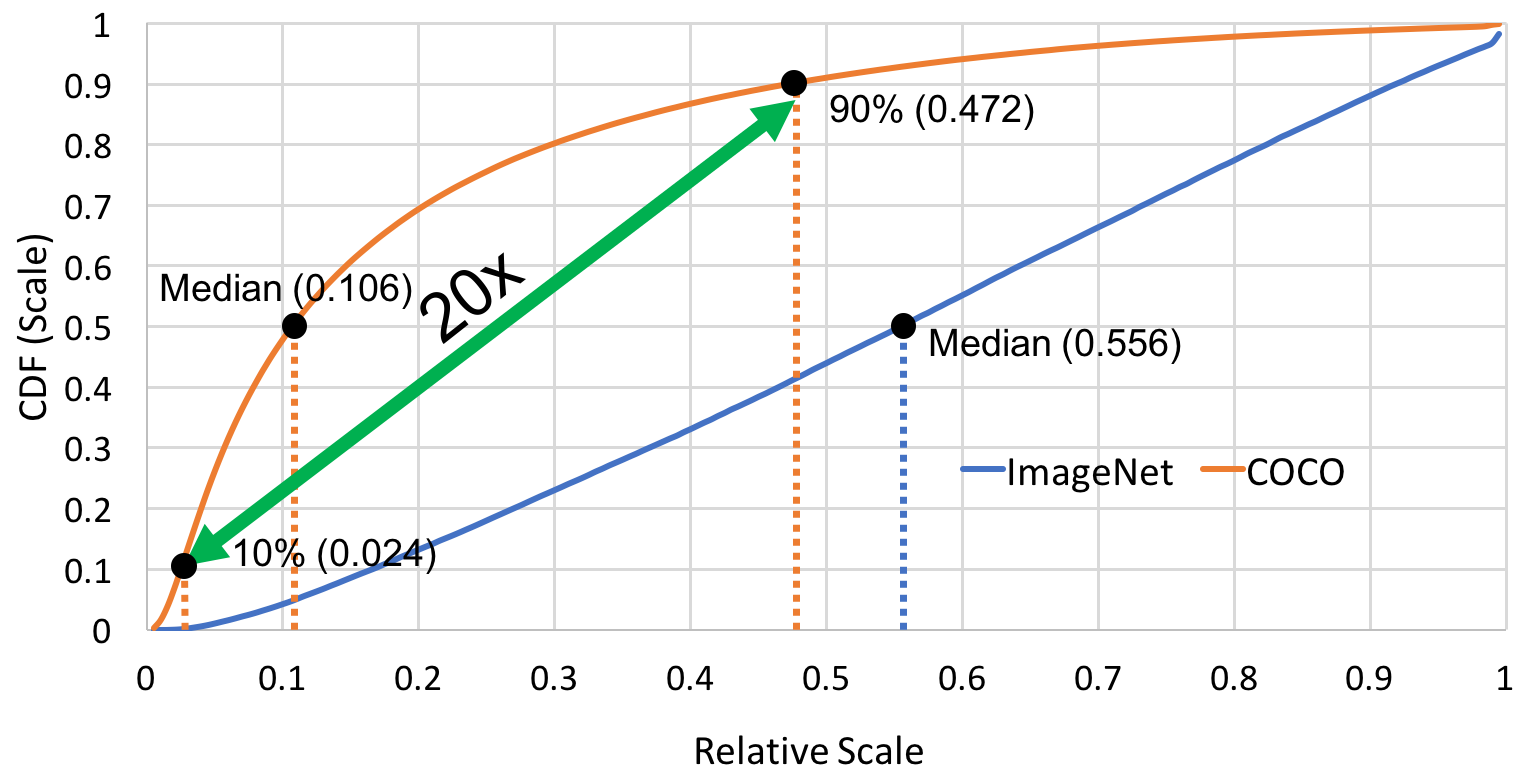}
    \caption{Fraction of RoIs in the dataset vs scale of RoIs relative to the image.}
    \label{fig:plot}
\end{figure}

\begin{figure*}
    \center
    \includegraphics[width=0.99\linewidth]{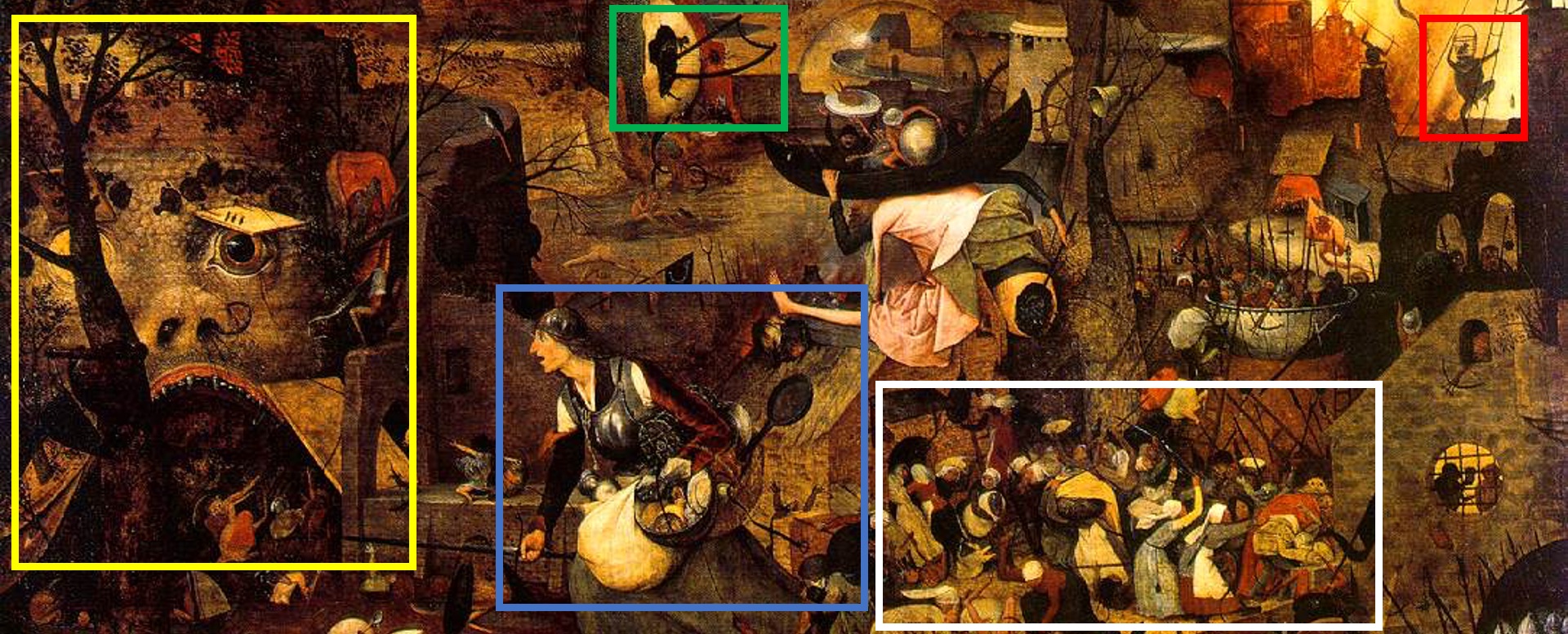}
    \caption{Dull Gret by Pieter Bruegel the Elder. In the center one can see Dull Gret, a brave lady who is taking an army of women with her (to her right) and she is invading hell (the entire picture is about hell). We can see devils at different places, for example on the left, top right or top center. Notice how difficult it is to focus on all objects in a complex scene at once. It is essential to {\em stop and focus} on different regions while skimming through the image to find all the objects.}
    \label{fig:bio_pic}
\end{figure*}

In order to motivate the challenges arising due to scale-variation, we invite the reader to take a look at the famous painting by the renaissance painter Pieter Bruegel, in Fig.~\ref{fig:bio_pic}. First, try to locate the objects shown in the image - there are far more objects than the overlaid bounding-boxes! If you have successfully detected more objects, you can appreciate the fact that differently scaled objects require careful attention to their intrinsic scales. Also, you would have noticed that you had to \emph{stop} and \emph{focus} on different regions while skimming through the image to find all the objects. Now, try focusing on any one object in the picture and observe the rest of the objects disappearing in the background \cite{corbetta2002control}. The phenomenon that you just witnessed is commonly referred to as \emph{foveal vision} \cite{rensink2000dynamic}, which affords adaptive shift and zoom for the object of interest in a scene.

Objects appearing in natural images also exhibit large variations in scale. To show this quantitatively, we plot the fraction of the size of an object in an image \vs its own size, from the COCO dataset \cite{lin2014microsoft}, in Fig. \ref{fig:plot}. From the figure, we notice that an object can appear at scales that differ by an order of magnitude - as was the case in the painting. Intuitively, such a wide scale-variation calls for object-detection frameworks that explicitly learn to tackle this challenge. Unfortunately, modern object detection frameworks are designed to perform inference at a single input resolution, or scale, to detect objects of all sizes \cite{redmon2016yolo9000,liu2016ssd,lin2017feature} and multi-scale inference is often shrugged under the {\em bells-and-whistles} category. This practice is in stark contrast to the scale-space theory \cite{lindeberg1993scale, lindeberg1998feature, witkin1984scale}, an extremely effective and theoretically sound framework to tackle naturally occurring scale-variation. The scale-space theory advocates the existence of a narrow range of scales at which an image structure can be optimally comprehended, much like a {\em fovea} in the scale-space. Since the foveal range of scale for an unknown image-structure is not available a priori, the image is convolved with multiple Gaussians and their derivatives, and the maximum activation is used to detect the intrinsic scale of the structure. Convolution by a filter bank of multi-scale Gaussian kernels or image-pyramids are popular mechanisms to implement the scale-space theory in practice.  This principle has been successfully applied to numerous problems throughout the history of computer vision \cite{canny1986computational, perona1990scale, lowe2004distinctive, lazebnik2006beyond, chen2013efficient} and such scale-invariant representations have been successfully employed for recognizing and localizing objects for problems like object detection, pose-estimation, instance segmentation \etc.

If the answer to scale variation in object-detection has been around for decades, why hasn't it been adopted to CNN-based object detectors? The answer lies in yet another critically important characteristic of real-world object-detection systems besides accuracy: \emph{computational cost}. Naturally, a multi-scale processing pipeline would increase the computational cost of already compute-hungry CNN systems that can potentially render them practically useless. To this end, we seek motivation from scale-space theory and the foveal nature of human-vision to propose a novel object detection paradigm that strikes a balance between accuracy and computational footprint. 

\begin{itemize}
    \item We carefully study the problem of object-detection under large scale-variation and provide crucial insights into the related challenges and their detrimental effects on current object-detection systems. We also discuss the possible shortcomings of the popular contemporary practices to tackle scale-variation, such as feature pyramids and multi-scale training/inference.
    \item After highlighting these shortcomings, we propose to re-scale all the objects during training and inference - just like human vision - to ensure their sizes range within a fixed interval only. Such a restriction of objects' size during training affords better learning of object-sensitive filters, and therefore, results in better accuracy. However, this re-scaling operation significantly increases the computational cost, both during training and inference stages, because it requires processing a multi-scale image pyramid (referred to as \emph{Scale Normalized Image Pyramid} or SNIP from now on for brevity).
    \item To address the increased computational cost during training, we propose an efficient \emph{spatial sub-sampling} scheme which only operates on fixed-size sub-regions likely to contain objects (as object locations are known during training). The resulting approach, referred to as \emph{Scale Normalized Image Pyramid with Efficient Resampling} or SNIPER, yields up to 3$\times$ speed-up during the training phase when used with SNIP. Unfortunately, during the inference phase we still need to processes the entire image pyramid as object locations are unknown.
    \item To address the computational cost during inference, we propose to process the image pyramid using a coarse-to-fine approach, and predict the locations and extent of object-like regions which get processed in successive scales. Intuitively, it's akin to active human-vision that first \emph{skims over} the field-of-view to spot interesting regions for further processing and only recognizes objects at the right resolution. The resulting algorithm is referred to as \emph{AutoFocus} and results in a 2.5-5$\times$ speed-up during inference when used with SNIP.
\end{itemize}

\section{Related Work} \label{sec:related}
In this section, we provide the details of methods that have tried to tackle similar challenges as our contribution and discuss the differences between our contributions and previous art. Since Multi-Scale representations are the major source of motivation for our work, we layout the detailed history of this approach before moving to the CNN-based multi-scale approaches.

\subsection{A Brief History of Multi-Scale Representations}
The vast amount of real-world visual information exhibits a meaningful semantic interpretation when analyzed within a task-dependent range of scales, or resolutions. This fact has been known to the researchers in computer vision for the past 50 years and has been employed for several vision problems since then. One of the earliest use of this idea came in the form of multi-scale operators for edge and curve detection \cite{rosenfeld1971Edge}, in 1971. Around the same time, a slightly different instance of multi-scale information processing was proposed in the form of recursive spatial sub-region processing at discrete spatial resolutions, more commonly known as \emph{quad-trees} ~\cite{klinger1971patterns}. A quad-tree based pattern-recognition approach, termed as \emph{recognition-cone} was developed in ~\cite{uhr1972layered}; similar ideas were explored in \cite{hanson1974processing, tanimoto1975hierarchical}. Eventually these ideas took the form of \emph{multi-scale image-pyramids} \cite{burt1981fast, crowley1981representation}, in 1981. Since then, multi-scale image-pyramids have served as a bedrock for numerous applications in the field of computer vision; ranging from simple edge/corner detection to complex object-detection systems \cite{canny1986computational, perona1990scale, lowe2004distinctive, lazebnik2006beyond, lin2017feature, chen2013efficient}.

While the early multi-scale image-pyramids enjoyed immense success in practical applications, a comprehensive theory of such representations came, in 1984, as the \emph{scale-space theory} that represents visual information in a continuous one-parameter scale-space \cite{witkin1984scale} formed by convolutions with Gaussian kernels at different scales. Such representations successively suppress the finer details without giving rise to new local minimas in the derived representation. The scale-space theory was further employed in the seminal work of Lindenberg \cite{lindeberg1998feature} for feature detection with automatic scale selection.

\subsection{CNN Based Multi-Scale Pyramids} \label{ssec:cnn_pyr}
Over the past few years, deep CNNs are used as the de-facto feature extraction algorithms for virtually all the computer-vision problems.  Intuitively, the layered representation of deep CNNs is akin to the multi-scale representations used in the past except that they are learned from the data. Moreover, the input is not the \emph{explicit} multi-scale image-pyramid, rather the intermediate representations of the deep CNNs are themselves used as a \emph{proxy for multi-scale presentations}. It is primarily done to save computational cost and these might fail to adapt to the immense variation in the scale-space of semantic information embedded in an image. Typically, these intermediate representations are at a resolution which is 32/64 times, depending on the backbone architecture, less than the original image. Therefore, in order to obtain sufficient resolution for small objects, common methods employ dilated/deformable convolutions \cite{chen2016deeplab,dai2017deformable}, or up-sampling the image by up to 4 times during inference \cite{dai2017deformable,dai2016r,he2016deep}. A few representative approaches try to improve object detection using intermediate feature representations, either by combining the multi-scale feature maps prior to detection \cite{bell2016inside, kong2016hypernet} or by making independent predictions at different layers to ensure that the smaller objects are trained at finer resolution layers (like conv3) while larger objects are trained at relatively coarser resolution layers (like conv5) \cite{cai2016unified, najibi2017ssh,yang2016exploit}. The success of these approaches hinges on the underlying, \emph{often unspoken}, assumption that smaller objects can be detected with a relatively less complex network and a more complex network is only needed for the large objects. Clearly, this is too strong an assumption that can hurt the performance for small objects. Ideally, we would want small objects to be processed at a high resolution with complex networks; the main motivation of our work, and the major contribution as well.

\begin{figure}[t]
\centering
\includegraphics[width=1\linewidth]{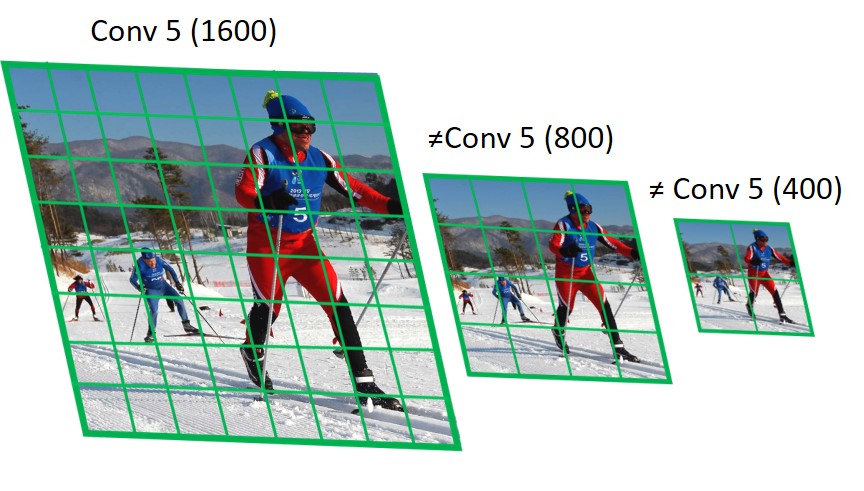}
\caption{The same layer convolutional features at different scales
of the image are different and map to different semantic regions in
the image at different scales.}
\label{fig:cnn_arch}
\end{figure} 

\subsection{Modifying CNN architectures \vs Image Pyramids}
Architectural changes to CNNs towards aggregation of multi-scale information to tackle scale-variation for detection \cite{lin2017feature,liu2018path,zhang2018single} are a popular technique for scale-invariant object-detection. Feature pyramid networks, FPN, \cite{lin2017feature} have shown impressive performance boost for detecting objects at different scales with a trivial increase in computation. Path Aggregation Network \cite{liu2018path}, PA-Net, aggregates multi-scale information in a an adaptive bottom-up manner to effectively fuse the information from the lower layers to the topmost feature layers. RefineDet \cite{zhang2018single} also combines features similar to feature pyramid networks to improve performance on objects of different scales. Such empirical success gives the impression that architectural changes are a comprehensive answer to the scale-variation challenge. However, the same layer convolutional features at different scales of the image are different and map to different semantic regions in the image at different scales, as shown in Figure \ref{fig:cnn_arch}. Our findings suggest that performing object-detection at a \emph{normalized} resolution is the key to address the problem of scale-variation. For example, consider images at a resolution of 4000$\times$3000 pixels (a typical smartphone camera resolution) that contain objects varying from 20 to 2000 pixels in size. If the detector is applied at a single scale only, it would require the same CNN filter to detect 20, 200, and 2000 pixel objects simultaneously! Naturally, learning such a detector is far more difficult than learning a detector that only needs to detect objects within a pre-defined narrow range of scales. 

\begin{figure*}[t]
    \centering
    \includegraphics[width=\linewidth]{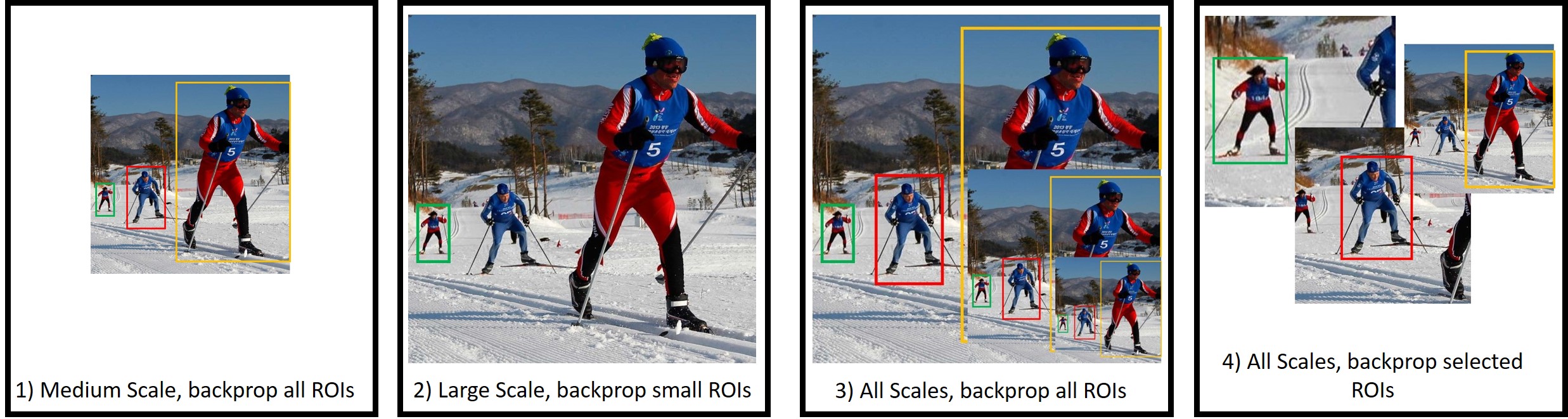}
    \caption{Different approaches for providing input for training the classifier of a proposal based detector.}
    \label{fig:mstr}
\end{figure*}

Furthermore, architectural changes (like in FPN/PA-Net/RefineDet) \emph{inherently} assume that combining less complex but spatially-dense network features (up to conv3) with more complex but spatially-sparse features are sufficient for detecting smaller objects. However, like \cite{cai2016unified, najibi2017ssh,yang2016exploit}, even here, lower layer filters are composed of only 3-4 convolution layers for contemporary backbone architectures. Therefore, they lack the sufficient complexity to cope up with large variations in pose, appearance, lighting, and occlusion, which occur, indiscriminately, for both the large as well as the small resolution objects. The deeper features are too spatially-sparse to represent individual parts of the objects, so they are not equivalent to extracting the features for a higher resolution image. Therefore, despite the embedded multi-scale processing, these methods still up-sample the input image. For example, in RetinaNet, which uses an FPN architecture, the mAP drops from 37.8\% to 31.9\% (a drop of 5.9\%) \cite{lin2017focal} when the input resolution is reduced from 800 pixels (shorter side) to the native image resolution of 400 pixels. If combining shallow and deep feature representations in different permutations and combinations was indeed the answer to scale-variation, this up-sampling step wouldn't be necessary. Note that when the image is up-sampled, the network is processing more of the easy regions in the image (like already large objects or background) and this increases quadratically with the up-sampling factor. Lastly, but importantly, up-sampling beyond a certain image resolution would also degrade the performance on large objects (for example at 4000$\times$3000) as the network may not have sufficient receptive field to represent the entire object.

An interesting anchor-free approach treats objects as points and learns to detect the bounding-box centers as key-points followed by regression on the height/width of the boxes \cite{zhou2019objects}. They leverage a cascaded bottom-up-bottom inference scheme to fuse multi-scale features to tackle scale-variation. Another anchor-free approach, \emph{FoveaBox}, uses a FPN backbone to first predict class-specific objectness maps followed by class-agnostic bounding-box predictions for object-detection \cite{kong2019foveabox}. Both \cite{zhou2019objects, kong2019foveabox}, fuse multi-scale features, similar to FPNs, therefore, suffer from similar sub-optimalities as discussed before.

Multi-scale image-pyramids explicitly up/down-sample object instances and, therefore, afford the employment of the same complexity convolution network to detect all the objects \emph{within a narrow scale-range} regardless of their original resolution. Hence, our framework employs multi-scale image pyramids to model the {\em stop} and {\em zoom} approach of the human-vision system. We explicitly take a multi-scale pyramid as the input but only process a small fraction of it by automatically selecting object-like regions at appropriate scales; much like the foveal vision in humans. It results in a more accurate system with reduced computational cost, especially when processing upsampled images in high resolutions. It is important to note that image pyramids applied on top of improved CNN feature representations still benefit from architectural improvements, therefore, our proposed approach is complementary to the aforementioned architectural improvements.

\subsection{Adoption of this work}
Since the publication of the conference versions \cite{singh2018SNIP, singh2018SNIPER, najibi2019autoFocus}, multiple works have extended and/or built on top of our findings to further improve the performance of object detection systems for different applications. Scale-aware TridentNet architecture learns scale-specific feature extractors by varying the size of receptive fields to marry the benefits of SNIPER and Feature Pyramid Networks \cite{lin2017feature}. Some application specific adoption of the ideas are presented for large-scale scale-invariant face detection \cite{Najibi_2019_CVPR}, large-scale object detection \cite{gao2018solution} and object-detection systems for aerial images \cite{yang2019clustered}. The success of image pyramid based methods is also evident in empirical comparisons in contemporary literature on object detection for high-resolution images and autonomous driving ~\cite{yang2019clustered,huang20201st}.

\section{Efficient Multi-Scale Object Detection}

This section presents our proposed multi-scale approach for object detection. In Sub-Sec.~\ref{subsec:object_det_behaviour}, we analyze the behavior of deep-CNN based detectors under scale variations to understand the inherent trade-offs involved in creating a multi-scale training pipeline. Based on these findings, we present scale-normalized image pyramids to tackle the large scale-variation in objects observed in 2D images in Sub-Sec.~\ref{subsec:snip}. Then, Sub-Sec.~\ref{subsec:sniper} presents the details of an approach (SNIPER) that affords training object detectors with scale-normalized image pyramids in an efficient fashion by using sub-regions of images instead of the entire image. Finally, we present an automatic region selection approach for efficient inference on scale-normalized image pyramids (AutoFocus) in Sec.~\ref{subsec:auto}.

\begin{figure*}
    \center
    \includegraphics[width=0.99\linewidth]{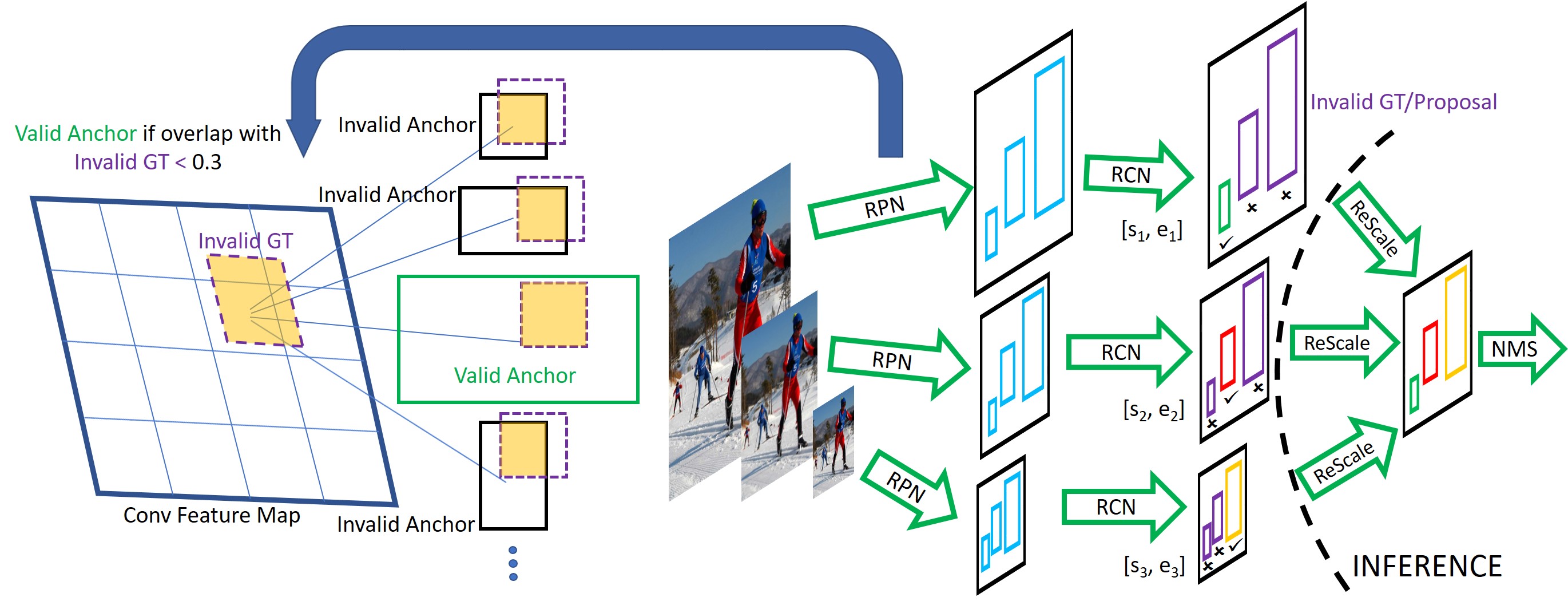}
    \caption{SNIP training and inference is shown. Invalid RoIs which fall outside the specified range at each scale are shown in purple. These are discarded during training and inference. Each batch during training consists of images sampled from a particular scale. Invalid GT boxes are used to invalidate anchors in RPN. Detections from each scale are rescaled and combined using NMS.}
    \label{fig:full}
\end{figure*}

\subsection{Disentangling Object Detection, Scale Variations and Deep-Learning Characteristics}\label{subsec:object_det_behaviour}
This section analyses the effect of image resolution, the scale of object instances, and variation in data on the performance of an object detector. We train detectors at different image resolutions and evaluate them on 1400x2000 images for detecting small objects (less than 32x32 pixels in the COCO dataset) only to tease apart the factors that affect the performance. The results are reported in Table \ref{tab:small}. 

\subsubsection{Training/Inference with large scale-variation} We start by training detectors (Deformable R-FCN \cite{dai2017deformable} in this particular example) that use all the object instances at two different pixel-resolutions, 800x1400 and 1400x2000, referred to as $800_{all}$ and $1400_{all}$, respectively, Fig \ref{fig:mstr}\textcolor{red}{.1}. As expected, $1400_{all}$ outperformed $800_{all}$, because the former is trained and tested on the same resolution \ie 1400x2000. However, the improvement is only marginal. Why? To answer this question we consider what happens to the medium-to-large object instances while training at such a large resolution. They become too big to be correctly classified! Therefore, training at higher resolutions scales up small objects for better classification, but blows up the medium-to-large objects which degrades performance.

\subsubsection{Scale specific detectors} Since the hypothesis was that large objects adversely affect the performance of small objects, this time we train another detector ($1400_{<80px}$) at a resolution of 1400x2000 while ignoring all the medium-to-large objects ($>80$ pixels, in the original image) to eliminate the deleterious-effects of extremely large objects, Fig \ref{fig:mstr}\textcolor{red}{.2}. Unfortunately, it performs significantly worse than even $800_{all}$. Why do we observe such a counter-intuitive result? It happens because we lose a significant source of variation in appearance and pose by ignoring medium-to-large objects (about 30\% of the total object instances) that hurts performance more than it helps by eliminating extreme-scale objects.

\subsubsection{Multi-Scale Training (MST)} So far, our experiments indicate that not only do we need to reduce scale-variation during training/inference, but we also require training data capturing diverse variations. Therefore, this time we evaluate the common practice of  randomly sampling images at multiple resolutions during training, referred to as MST - as was the case in Fast-RCNN \cite{girshick2015fast} \footnote{MST also uses a resolution of 480x800}, Fig \ref{fig:mstr}\textcolor{red}{.3}. It ensures that training instances are observed at many different resolutions, but its performance is degraded by extremely small and large objects. Consequently, it performed similarly to $800_{all}$. Effectively, the benefits of observing all object instances at all scales are offset by extremely large/small objects that force the network to learn object-sensitive filters across a wide range of scales. Therefore, we conclude that it is important to train a detector with appropriately scaled objects while capturing as much variation across the objects as possible. In the next section, we describe our proposed solution that achieves exactly this and show that it outperforms current training pipelines.

\begin{table}[ht!]
    \centering
    \caption{mAP on Small Objects (smaller than 32x32 pixels) under different training protocols. MST denotes multi-scale training as shown in Fig. \ref{fig:mstr}\textcolor{red}{.3}. R-FCN detector with ResNet-50 (see Section 4).}
    \label{tab:small}
    \begin{tabular}{|c|c|c|c|c|}
        \hline
          1400$_{<80px}$  & \hspace{0.04cm} 800$_{all}$ \hspace{0.04cm} & 1400$_{all}$ & \hspace{0.1cm} MST \hspace{0.1cm} & \hspace{0.08cm} SNIP \hspace{0.08cm} \\
        \toprule
        16.4 & 19.6 & 19.9 & 19.5 & 21.4 \\

        \bottomrule
    \end{tabular}

\end{table}


\subsection{Scale Normalization for Image Pyramids}
\label{subsec:snip}

In the scale-space theory, the width of the Gaussian kernels is varied to achieve the maximum response for an image structure in the scale space. For CNN-based object-detection systems, the {\em deep filter} is fixed in the form of a learnable convolutional neural network which comprises of millions of parameters. There does not exist a parameter (as was the case with the Gaussian kernel) which can be tuned to obtain normalized responses for different sized structures in the image. However, the processing, or filtering, of an image with this deep filter results in a convolutional feature map whose resolution can be computed from the image resolution via a known mapping (depending on the stride of the network). Therefore, the extent of an object in this feature map depends on its pixel resolution in the input image, through the aforementioned mapping. Hence, we can control the extent/size of any object in the feature map by appropriately re-scaling the input image that contains the object. This ensures that even extremely different-sized objects are mapped to a similar-sized projection in the feature map via appropriate scaling of the input image. In such a formulation, the rich complexity of the deep-network can be employed for both small and large objects while effectively maintaining similar projection sizes in the feature map for both. This affords the learning of detectors with a narrow range of scale variations in the feature map.

Now that we can control the size of object projections in the feature map, we turn our attention to deciding the optimal size of such projections. On one hand, an object's projection must span a minimum spatial resolution in the feature map to have sufficient information to recognize it. On the other hand, such projections should have sufficient context to capture relationships across parts of an object to aid its classification and shouldn't be packed too densely to avoid confusion. Hence, we propose to ensure that each object projection spans between 5 to 15 spatial points in the feature map. Based on our experiments, we found this range to work the best. Using this guideline, for any CNN we can decide the size of objects to train for each resolution in a multi-scale image pyramid. Thereby, achieving the goal of attending to each object at the appropriate resolution. To this end, we propose a modified version of MST where only the object instances falling within a pre-defined scale range (or min/max resolution) are used for training the detector.

Training the detector with this simple modification takes care of the biggest drawback of MST where each object instance was observed at every resolution. In MST, at a high resolution (like 1400x2000) large objects were hard to classify and at a low resolution (like 480x800), small objects. Fortunately, each object instance appears at several different resolutions and some of those appearances fall in the pre-defined scale range where we get the maximum response from our {\em deep filter}. Hence, training is only performed on objects that fall in the pre-defined scale range and the remainder is simply ignored during back-propagation. We refer to this representation as \emph{Scale Normalization for Image Pyramids} or SNIP. SNIP ends up employing all the object instances during training, which helps capture all the variations in appearance and pose, while reducing the large scale-space variation. This also helps to make the task of CNNs easier because now they only need to learn object-sensitive filters for a small range of scale variations. The result of evaluating the detector trained using SNIP is reported in Table \ref{tab:small}. We see that it outperforms all the other approaches for multi-scale training. This experiment demonstrates the effectiveness of SNIP for object-detection. Below we discuss the implementation of SNIP in detail.

\subsubsection{Training with SNIP}\label{sssec:snip_training}
For training the classifier-head of the object detector, all ground-truth boxes are used to assign labels to proposals. But, we do not select the proposals and ground truth boxes that are outside the pre-defined scale-range during training. We use the pixel area as a measure to decide whether a proposal or ground-truth box falls within the desired scale. At a particular image resolution $i$, if the area of an RoI $ar(r)$ falls within a range $[s_i,e_i]$, it is marked as \emph{valid}, else it is \emph{invalid}. The label $l(r)$ for any RoI is defined as follows, 

\[
    l(r) = 
\begin{cases}
    l_{GT},& IoU(GT, r) >= 0.5, s_i < ar(r) < e_i \\
    0,& IoU(GT, l) < 0.5, s_i < ar(r) < e_i  \\
    -1,& ar(r) <= s_i \\
    -1,& ar(r) >= e_i  \\
\end{cases}
\]

where, $IoU(GT, r)$ is the Intersection-Over-Union score between the ground-truth and an RoI and $l_{GT}$ is the label of the ground-truth box. To ensure all the data is used, training samples still falling outside the range are included when processing the smallest/largest resolution. Once all the ground-truth boxes are marked either valid or invalid, a similar idea is used for training the region-proposal network as well. First, all the ground-truth boxes are used to assign labels to anchors. Then, the anchors that have an overlap greater than 0.3 with an \emph{invalid ground-truth box} are excluded during training (\ie their gradients are set to zero). 

\subsubsection{Inference with SNIP}\label{snip_inference}
During inference, we generate proposals using RPN for each image-resolution and classify them independently at each resolution as well, as shown in Fig \ref{fig:full}. Similar to the training phase, we do not select detections (not proposals) which fall outside a specified range at each resolution. After classification and bounding-box regression, we use soft-NMS \cite{bodla2017} to combine detections from multiple resolutions to obtain the final detection boxes, refer to Fig. \ref{fig:full}. While SNIP does an excellent job at capturing the training-data variability within a restricted scale-range, it comes with an additional computation cost both during training and inference. Moreover, processing multi-scale image pyramids also increases the memory requirements per batch that can be prohibitive for training on smaller-memory GPU cards. Therefore, in the next section, we propose an efficient spatial sub-sampling algorithm to reduce the computational cost during training.
\subsection{Scale-Normalized Image Pyramids with Efficient Resampling or SNIPER}
\label{subsec:sniper}

The proposed SNIP framework requires the processing of a multi-scale image pyramid that can result in images of sizes up to 1400$\times$2000 pixels. Training on such high-resolution images with deep networks, like ResNet-101 requires prohibitively large GPU memory and significantly increases the computational budget. We propose to restrict the input size to a reasonable pixel-resolution while still covering all the object instances. This sub-section presents the details of the proposed strategy to select sub-regions (or chips) from the multi-scale image pyramids for efficient and accurate training of object detectors. 

\subsubsection{Efficient Chip-Generation with Image Content Layout}
Fortunately, we can exploit the characteristics of the SNIP training along with the semantic layout of object instances to only use sub-regions (or chips) for training object detectors. To build the intuition, let's start by considering how small objects are treated in SNIP. By design, SNIP is likely to ignore the gradients coming for extreme-scale object instances, hence, at a $3\times$ resolution, it will mostly ignore the medium/large-sized objects and only attend to the objects that were small in the original image resolution. Therefore, we do not need to process the entire image at $3\times$ resolution and it's sufficient to just sample multiple small-sized chips around the \emph{originally} small objects at $3\times$ resolution. Now, let's focus our attention on what happens with the \emph{originally} large-sized objects. Well, they become even larger at higher-resolution and would definitely get ignored during training. Hence, there isn't any benefit of up-sampling if the original image is already high-resolution and only contains large-size objects. The above discussion suggests that perhaps we can simply ignore large portions of images that do not contain objects within the desired scale-range and only choose multiple small-size chips at different resolutions that \emph{tightly} cover the objects of interest. Such a spatial sub-sampling would indeed save a lot of computation, but, it will come at a heavy cost of losing contextual information, which has been proved to be critical for accurate recognition \cite{torralba2003contextual,divvala2009empirical,yu2015multi}. Moreover, it will also remove a significant portion of the background at higher resolutions, which will lead to a biased training data-distribution in favor of foreground regions, which can negatively affect the training \cite{uijlings2013selective,girshick2014rich}. Therefore, we face a trade-off between computation, context, and negative mining while trying to accelerate multi-scale training.

To this end, we propose a sampling strategy, referred to as \textit{Scale Normalization for Image Pyramids with Efficient Re-sampling (SNIPER)}, which adaptively samples chips from multiple scales of an image pyramid, conditioned on the image content. The positive chips are conditioned on the ground-truth instances and negative chips are based on proposals generated by a region proposal network. It affords to reduce the image resolution from $1400\times2000$ pixels to $512\times512$ pixel chips, which, in turn, leads to an overall reduction of $3\times$ in training time. Moreover, $512$x$512$ pixel chips can be trained with a large batch-size with batch-normalization on a single GPU node. In particular, we can use a batch size of $20$ per GPU (leading to a total batch size of 160), even with a ResNet-101 based Faster-RCNN detector. Formally, SNIPER generates chips $\mathcal{C}^i$ at multiple scales $\{s_1, s_2,.., s_i,.. s_n\}$ in the image. For each scale, the image is first re-sized to width ($W_i$) and height ($H_i$). On this canvas, $K \times K$ pixel chips are placed at equal intervals of $d$ pixels (we set $d$ to $32$ in this paper). This leads to a two-dimensional array of chips at each scale, $\mathcal{L}^i$. In the following sections, we describe the details of positive and negative chip-generation and the label assignment process.

\begin{figure}[t]
\centering
\includegraphics[width=0.8\linewidth]{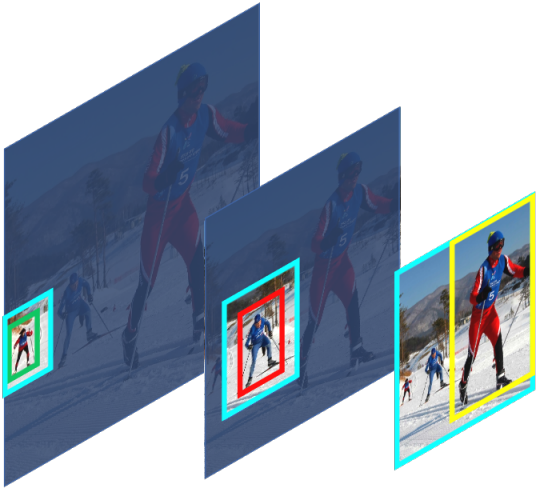}
\caption{SNIPER selectively trains the detector only on parts of the input pyramid (chips) while respecting the SNIP ranges. The light blue rectangle show the selected chips in each pyramid scale. The valid objects in each chip is also shown with colored boxes.}
\label{fig:pos_chips}
\end{figure}

\begin{algorithm}[h!]
    \caption{SNIPER: Positive Training Chips Generation}
        \label{alg:sniper}
	\SetKwInOut{Input}{Input}\SetKwInOut{Output}{Output}
    \Input{Image: $I$, Ground-Truth: $\mathcal{B}$, Image-Resolutions: $\mathcal{S} = \{(W^1, H^1) \dots (W^n,H^n)\}$, chip-size: $K$, stride: $d$, valid ROI-area ranges: $\mathcal{R} = \{[r_{min}^1, r_{max}^1] \dots [r_{min}^n, r_{max}^n]\}$ } 
    \Output{Positive Chips: $\mathcal{C}_{pos}$}
    \BlankLine
        $\mathcal{C}_{pos} \gets \emptyset$ \\
        \For{$i \in \{1 \dots n\}$}{
        $I^i \gets $ imresize$(I,(W^i,H^i))$ //resize original image\\
        $\mathcal{B}^i \gets $ resize$(\mathcal{B},(W^i,H^i))$ //resize g.t. boxes\\
        $\mathcal{G}^i\gets \mathcal{B}^i \in [r_{min}^i, r_{max}^i]$ //select valid size g.t. boxes \\
        $\mathcal{L}^i \gets K \times K$ chips from $I^i$; $stride=d$  //all chips \\
        $\mathcal{C}^i_{pos} \gets \emptyset$ \\
        
        \While{$\mathcal{G}^i \neq \emptyset$}
        {
        $\mathcal{L}^i_{max} \gets $ chip covering max number of g.t. boxes \\
        $\mathcal{G}^i \gets \mathcal{G}^i$ - covered g.t. boxes \\
        $\mathcal{C}^i_{pos}\gets \mathcal{C}^i_{pos} + \mathcal{L}^i_{max}$;  $\mathcal{L}^i \gets \mathcal{L}^i - \mathcal{L}^i_{max}$\\
        }
        $\mathcal{C}_{pos} \gets \mathcal{C}_{pos} + \mathcal{C}^i_{pos}$ \\
        }
 	\Return{Chips $\mathcal{C}_{pos}$}
 	
\end{algorithm}

\begin{figure*}
\centering
\includegraphics[width=\linewidth]{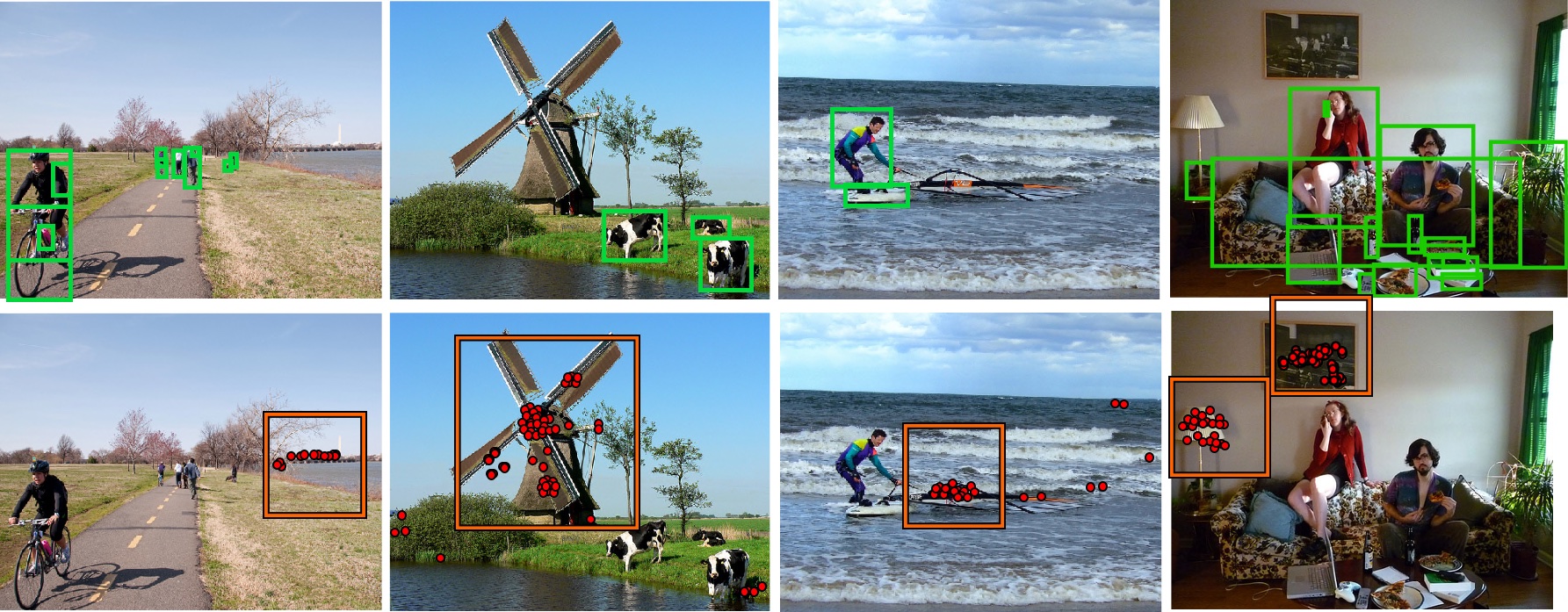}
\caption{SNIPER negative chip selection. First row: the image and the ground-truth boxes. Bottom row: negative proposals not covered in positive chips (represented by red circles located at the center of each proposal for the clarity) and the generated negative chips based on the proposals (represented by orange rectangles).}
\label{fig:neg_chips}
\end{figure*}

\subsubsection{Positive Chip Selection}\label{sssec:sniper_pos}


For each scale, there is a desired range of RoI area $\mathcal{R}^i = [r_{min}^{i}, r_{max}^{i}]$, $i \in [1, n]$ that determines the valid ground-truth boxes/proposals for training at each scale $i$. The valid list of ground-truth bounding boxes that lie in $\mathcal{R}^i$ is referred to as $\mathcal{G}^i$. The training chips are greedily selected, in a sequential manner, to cover the maximum number of valid ground-truth boxes ($\mathcal{G}^i$) while minimizing the required number of chips. A ground-truth box is said to be covered if it is completely enclosed inside a chip. The set of all the positive chips from the $i^{th}$ scale are combined per image and are referred to as $\mathcal{C}_{pos}^i$. Since consecutive $\mathcal{R}^i$ contain overlapping intervals, a ground-truth box may be assigned to multiple chips at different scales. It is also possible that the same ground-truth box may be covered in multiple chips at the same scale. Ground-truth instances that have a partial overlap (IoU $>$ 0) with a chip are cropped. All the cropped ground-truth boxes (valid or invalid) are retained in the chip and are used in label assignment. This process is described in Algorithm \ref{alg:sniper}.

Thus, every ground-truth box is covered at the appropriate scale. Since the crop-size is much smaller than the resolution of the image (\ie more than $10\times$ smaller for high-resolution images), SNIPER does not process most of the background at high-resolutions. This leads to significant savings in compute and memory footprint while processing high-resolution images. In order to illustrate the above process, we use an example shown in Figure \ref{fig:pos_chips}. We show an image pyramid with valid ground-truth boxes in green, red and yellow and the generated positive chips by SNIPER in light blue. We can see that all the ground-truth boxes are covered in one of the generated chips. This is how SNIPER efficiently processes all ground-truth objects at an appropriate scale by forming low-resolution chips, corresponding to properly scaled objects from different scales.

\subsubsection{Negative Chip Selection}\label{sssec:sniper_neg}
Since the positive chips are optimized to cover the objects, they don't capture sufficient background to create a balanced sampling of positive and negative regions. Therefore, the resulting classifier can wrongly classify a lot of background as object-regions, leading to a high false positive rate. In order to tackle this challenge, contemporary object detection algorithms that use multi-scale training, use the entire image at all scales. Although training on all scales reduces the false positive rate, it also increases computation. We, on the other hand, posit that a significant amount of the background is \emph{relatively} easy to classify, and therefore, can be removed from training to save computation. In order to realize this intuition, we employ object proposals to identify the regions where objects are likely to be present. After all, our classifier operates on region proposals and the parts of image without any region proposals are easy-to-identify background and can be safely ignored during training.

Hence, for negative-chip mining, we first train the RPN for a couple of epochs without using negative-chips for training. Since the task of this network is to roughly indicate the regions that may contain false positives, it's not necessary for it to be very accurate. We use this RPN to generate proposals over the entire training set and record the portions in images that don't contain any proposals as \emph{easy backgrounds}. Now, for negative chip selection, at each scale $i$, we first remove all the proposals covered in $\mathcal{C}_{pos}^i$. Then, at each scale $i$, we greedily select all the chips that cover at least $M$ proposals in $\mathcal{R}^i$. This generates a set of negative-chips for each scale per image, $\mathcal{C}_{neg}^i$. During training, we randomly sample a fixed number of negative chips per epoch (per image) from this pool of negative-chips at all scales, \ie $\bigcup_{i=1}^{n} \mathcal{C}_{neg}^i$. Figure \ref{fig:neg_chips} shows examples of the generated negative chips by SNIPER. The first row shows the image and the ground-truth boxes, the bottom shows the proposals that are not covered by $\mathcal{C}_{pos}^i$ and the corresponding negative-chips (the orange boxes). For clarity, we represent each proposal by a red circle in its center. We can see that SNIPER only processes regions which can likely contain false positives, while safely ignoring large portions of the image, which leads to faster processing time. Intuitively, this process is akin to hard-negative mining for those chips that contain difficult background regions.

\subsubsection{Label Assignment}
Once the negative and positive chips are selected, our network is trained end-to-end on these chips like Faster-RCNN, \ie it learns to generate proposals as well as classify them with a single network. While training, proposals generated by RPN are assigned labels and bounding box targets (for regression) based on {\em all} the ground-truth boxes that are present inside the chip. We do not filter ground-truth boxes based on $\mathcal{R}^i$. Instead, the proposals that don't fall in $\mathcal{R}^i$ are ignored during training, or their gradients are not back-propagated. Therefore, a large ground-truth box that is cropped, can generate a valid proposal that is small. Like Fast-RCNN, we assign a positive label and bounding-box targets to all the proposals that have an overlap greater than 0.5 with a ground-truth box. Our network is trained end-to-end and we generate 300 proposals per chip. We do not constraint any fraction of these proposals for re-sampling as positives \cite{ren2015faster}, as it's done in Fast-RCNN. We did not use OHEM \cite{shrivastava2016training} for classification, instead, we use a simple softmax cross-entropy loss for classification. For assigning the RPN labels, we use valid ground-truth boxes to assign labels and invalid ground-truth boxes to invalidate anchors, as it's done in SNIP.

\subsubsection{Benefits}
For training, we randomly sample chips from the whole dataset for generating a batch. On average, we generate $\sim 5$ chips of size $512$x$512$ per image on the COCO dataset (including negative chips) when training on three scales ($512$/ms \footnote{max($width_{im}$,$height_{im}$)}, $1.667$, $3$). This is only $30$\% more than the number of pixels processed per image when single scale training is performed with an image resolution of $800$x$1333$. Since all our images are of the same size, data is much better packed leading to better GPU utilization which easily overcomes the extra $30$\% overhead. But more importantly, {\em we reap the benefits of multi-scale training on $3$ scales, large batch size and training with batch-normalization without any slowdown in performance on a single 8 GPU node!}.

It is commonly believed that high-resolution images (\eg $800$x$1333$) are necessary for instance-level recognition tasks. Therefore, for instance-level recognition tasks, it was not possible to train with batch-normalization statistics computed on a single GPU. Methods like synchronized batch-normalization \cite{liu2018path, zhao2017pyramid} or training on $128$ GPUs \cite{peng2017megdet} have been proposed to alleviate this problem. Synchronized batch-normalization slows down training significantly and training on $128$ GPUs is also impractical for most people. Therefore, group normalization \cite{GroupNorm2018} has been recently proposed so that instance-level recognition tasks can benefit from another form of normalization in a low batch setting during training. With SNIPER, we show that the image resolution bottleneck can be alleviated for instance-level recognition tasks. As long as we can cover negatives and use appropriate scale normalization methods, we can train with a large batch size of resampled low-resolution chips, even on challenging datasets like COCO. Our results suggest that context beyond a certain field of view may not be beneficial during training. It is also possible that the effective receptive field of deep neural networks is not large enough to leverage far away pixels in the image, as suggested in \cite{luo2016understanding}.


\begin{figure}
\centering
\includegraphics[width=0.82\linewidth]{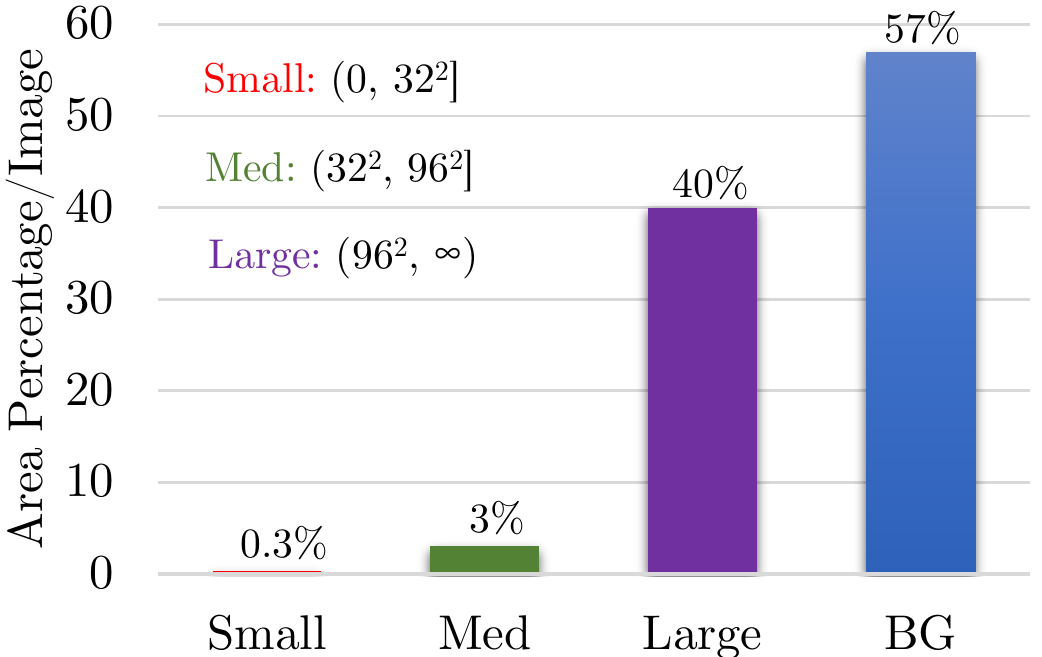}
\caption{Area of objects of different sizes and the background in the COCO validation set. Objects are divided based on their area (in pixels) into small, medium, and large.}
\label{fig:teaser_auto}
\end{figure}

\begin{figure*}
\centering
    \includegraphics[width=0.88\textwidth]{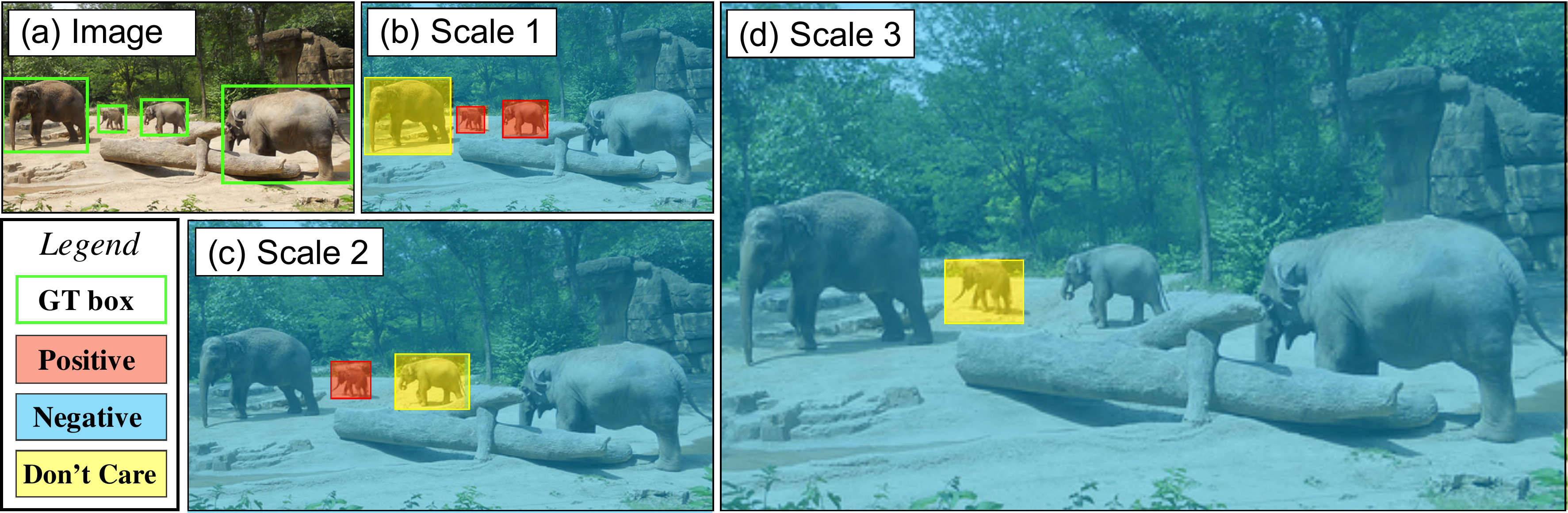}
\caption{The figure illustrates how FocusPixels are assigned at multiple scales of an image. At scale 1 (b), the smallest two elephants generate FocusPixels, the largest one is marked as background and the one on the left is ignored during training to avoid penalizing the network for borderline cases (see Sec. \ref{sec:focus_pixels} for assignment details). The labelling changes at scales 2 and 3 as the objects occupy more pixels. For example, only the smallest elephant would generate FocusPixels at scale 2 and the largest two elephants would generate negative labels.}
\label{fig:introPics}
\end{figure*}
\subsection{AutoFocus}
\label{subsec:auto}

While multi-scale processing brings significant improvements in accuracy, it comes at a computational cost, especially during inference. This is because the CNN is applied on all scales without factoring the spatial layout of the scene. To provide some perspective, we show the percentage of pixels occupied per image for different size objects in the COCO dataset in Fig \ref{fig:teaser_auto}. Even though 40\% of the object instances are small, they only occupy 0.3\% of the area. If the image pyramid includes a scale of 3, then just to detect such a small fraction of the dataset, we end up performing 9 times more computation at finer-scales. If we add some padding around small objects to provide spatial context and only upsample these regions, their area would still be small compared to the resolution of the original image. So, when performing multi-scale inference, can we predict regions containing small objects from coarser scales?

So far, we have exploited the semantic layout of training images and ground-truth bounding-boxes to efficiently sample medium-sized chips, $512\times512$ pixels, from a multi-scale image pyramid to constrain the scale-range of training object instances and reduce computation. Unfortunately, this technique is not applicable to the inference stage due to the lack of ground-truth information, which leads to large inference-time computation. Fortunately, we are not the first one to come across this problem and previous work has dealt with similar problems. Specifically, hand-crafted gradient-based features like SIFT \cite{lowe2004distinctive} or SURF \cite{bay2006surf}, combine two major components - the detector and the descriptor. The detector typically involves lightweight operators like Difference of Gaussians (DoG) \cite{marr1980theory}, Harris Affine \cite{mikolajczyk2004scale}, Laplacian of Gaussians (LoG) \cite{burt1987laplacian} \etc. and is applied to the complete image for finding {\em interesting} regions. The computationally heavy descriptor is only applied to interesting regions. Such cascaded processing of the image makes the entire pipeline computationally efficient. 

We seek motivation from the aforementioned cascaded systems, and propose a novel framework that first processes the coarsest scale and predicts the interesting regions in the image at the next scale. It continues processing the finer-level scales, or higher resolutions, in a sequential manner and keeps predicting interesting regions at the next scale until the entire pyramid is not processed. It re-scales and crops only the detected interesting regions for applying compute-heavy detectors. AutoFocus is comprised of three main components: the first learns to predict {\em FocusPixels}, the second generates {\em FocusChips} for efficient inference and the third merges detections from multiple scales, which we refer to as {\em focus stacking} for object detection. The details of each of the components are described in the subsequent sections. 

\subsubsection{FocusPixels}
\label{sec:focus_pixels}
FocusPixels are defined at the granularity of the convolutional feature map (like conv5). A pixel in a feature map is labeled as a FocusPixel if it has \emph{any} overlap with a small object. An object is considered small if it falls within a pre-defined area range (between $5\times5$ and $64\times64$ pixels in our implementation) in a re-sized chip (Sec. \ref{sec:chips}). During training, FocusPixels are marked as positives. Pixels that overlap with objects even smaller than the small objects, $\le 5\times5$ pixels are marked invalid. It's because such objects become even smaller after down-sampling and the network doesn't have sufficient information to predict their location at the next scale. We also mark the pixels that overlap with objects whose sizes range between $64\times64$ and $90\times90$ as invalid. It's due to the fact that the transition from small to large objects doesn't have a sharp boundary in terms of size. The rest of the feature-map pixels are marked as negative. AutoFocus is trained to generate high-value activations in the regions that contain FocusPixels.

\begin{figure*}[!ht]
\centering
\includegraphics[width=\linewidth]{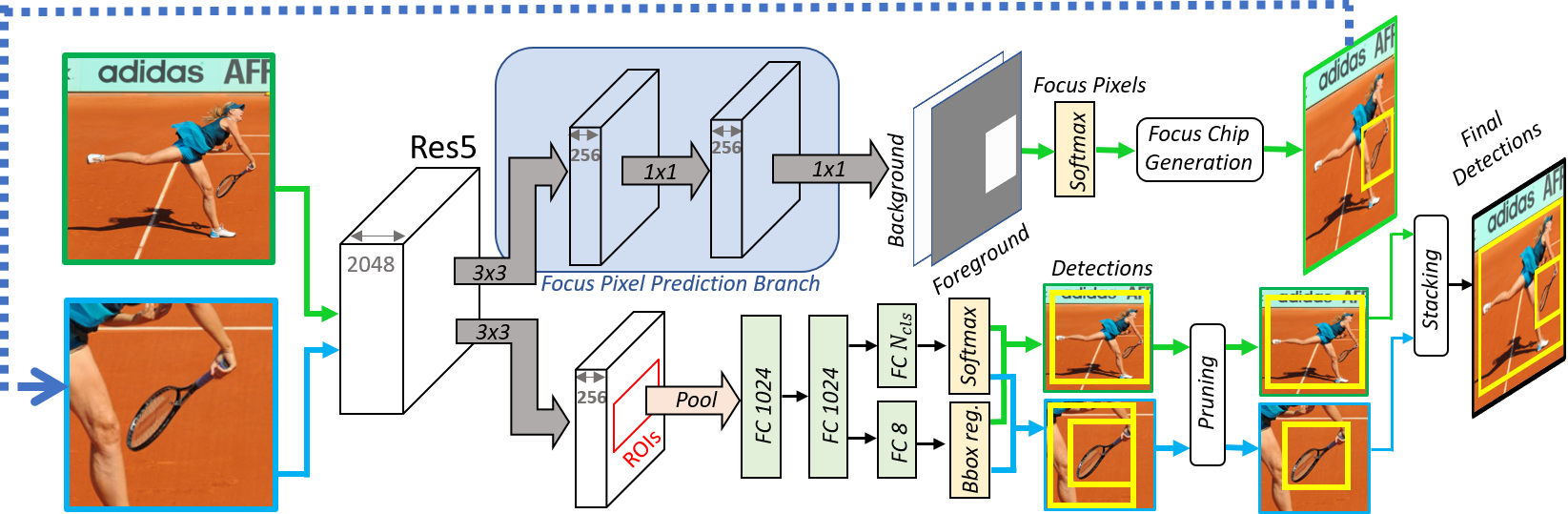}
\caption{The figure illustrates how AutoFocus detects a person and a racket in an image. The green borders and arrows are for inference at the original resolution. The blue borders and arrows are shown when inference is performed inside FocusChips. In the first iteration, the network detects the person and also generates a heat-map to mark regions containing small objects. This is depicted in the white/grey map - it is used to generate FocusChips. In the next iteration, the detector is then applied inside FocusChips only. Inside FocusChips, there could be detections for the cropped object present at the larger resolution. Such detections are pruned and finally, valid detections are stacked across multiple scales.}
\label{fig:framework}
\end{figure*}

Formally, for an image of size $X \times Y$, and a fully convolutional neural network with stride $s$, the resulting labels $L$ will be of size $X' \times Y'$, where $X' = \lceil \frac{X}{s} \rceil$ and $Y' = \lceil \frac{Y}{s} \rceil$. Since the stride is $s$, each label $l \in L$ corresponds to $s \times s$ pixels in the image. The label $l$ is defined as follows,

\[
    l = 
\begin{cases}
    1,& IoU(GT, l) > 0, a < \sqrt{GTArea} < b \\
    -1,& IoU(GT, l) > 0, \sqrt{GTArea} < a  \\
    -1,& IoU(GT, l) > 0, b < \sqrt{GTArea} < c  \\
    0,              & \text{otherwise}
\end{cases}
\]
where $IoU$ is the Intersection-Over-Union score of the $s\times s$ label block with the ground-truth bounding box, $GTArea$ is the area of the re-scaled ground-truth bounding box, $a$ is typically 5, $b$ is 64, and $c$ is 90. If multiple ground-truth bounding boxes overlap with a pixel, FocusPixels ($l=1$) are given precedence. Since our network is trained on 512 $\times$ 512 pixel chips, the ratio between the positive and negative pixels is around 10, so we do not perform any re-weighting for the loss. Note that during multi-scale training, the same ground-truth could generate a label of 1, 0 or -1 depending on how much it has been scaled. The labeling scheme is visually depicted in Fig \ref{fig:introPics}. For training the network, we add two convolutional layers (3$\times$3 and 1$\times$1) with ReLU non-linearity on top of the conv5 feature-map. Finally, we have a binary softmax classifier to predict FocusPixels, shown in Fig \ref{fig:framework}.

\begin{figure*}[tp]
\centering
\includegraphics[width=1\linewidth]{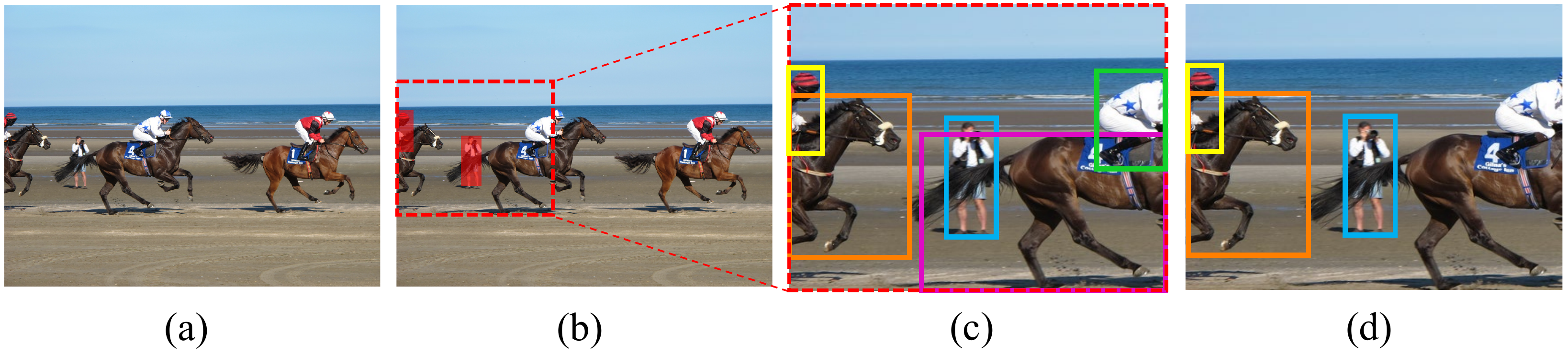}
\caption{Pruning detections while FocusStacking. (a) Original Image (b) The predicted FocusPixels and the generated FocusChip (c) Detection output by the network (d) Final detections for the FocusChip after pruning.}
\label{fig:FocusStacking}
\end{figure*}

\subsubsection{FocusChip Generation}
\label{sec:chips}
Armed with the capability of estimating the foreground probability at every pixel, we now turn our attention to obtain rectangular sub-regions, or FocusChips, for further processing with a CNN. During inference, we use a parameter, $t$, to mark the pixels, $\mathcal{P}$, whose foreground probability is greater than $t$ as FocusPixels. Consequently, a higher value of $t$ will lead to a smaller number of FocusPixel for further processing. Therefore, $t$ controls the speed-up and can be set with respect to the desired speed-accuracy trade-off. The thresholding with $t$ generates a set of connected components $\mathcal{S}$, which are dilated with a $d \times d$ sized-filter to increase the amount of required contextual information for recognition. As a result of dilation, previously disconnected components can form a new connection. Such components are merged to obtain the final set of connected components. Finally, we generate chips $\mathcal{C}$ that enclose the set of the aforementioned connected components. Note that the chips containing two connected components could overlap. As a result, these chips are merged with each other and replaced with their enclosing bounding-boxes in $\mathcal{C}$. Some connected components could be extremely small, and potentially lack the required contextual information for accurate recognition. Many small chips also increase fragmentation which results in a wide range of chip sizes. This makes batch-inference inefficient. To avoid these problems, we ensure that the height and width of a chip is greater than a minimum size $k$.  This process is described in Algorithm \ref{alg:chip_generation}. With the help of the identified FocusChips, we perform multi-scale inference on an image pyramid while focusing on regions that are more likely to contain objects.

\begin{algorithm}[t!]
    \caption{FocusChip Generator}
        \label{alg:chip_generation}
	\SetKwInOut{Input}{Input}\SetKwInOut{Output}{Output}
    \Input{Predictions for feature map $\mathcal{P}$, threshold $t$, dilation constant $d$, minimum size of chip $k$} 
    \Output{Chips $\mathcal{C}$}
    \BlankLine
        Transform $\mathcal{P}$ into a binary map using the threshold $t$ \\
        Dilate $\mathcal{P}$ with a $d\times d$ filter\\
        Obtain a set of connected components $\mathcal{S}$ from $\mathcal{P}$\\
        Generate enclosing chips $\mathcal{C}$ of size $> k$ for each component in $\mathcal{S}$\\
        Merge chips $\mathcal{C}$ if they overlap\\
 	\Return{Chips $\mathcal{C}$}
 	
\end{algorithm}

\subsubsection{Focus Stacking for Object Detection}
One issue with such cascaded multi-scale inference is that some detections at the boundary of the chips can be generated for cropped objects which were originally large. At the next scale, due to cropping, they could become small and generate false positives, such as the detections for the horse and the horse rider on the right, shown in Fig \ref{fig:FocusStacking} \textcolor{red}{(c)}. To alleviate this effect, Step 2 in Algorithm \ref{alg:chip_generation} is very important. Note that when we dilate the map $\mathcal{P}$ and generate chips, this ensures that no {\em interesting} object at the next scale would be observed at the boundaries of the chip (unless the chip shares a border with the image boundary). Otherwise, it would be enclosed by the chip, as these are generated around the dilated maps. Therefore, if a detection in the zoomed-in chip is observed at the boundary, we discard it, even if it is within valid SNIP ranges, such as the horse rider eliminated in Fig \ref{fig:FocusStacking} \textcolor{red}{(d)}. 

There are some corner cases when the detection is at the boundary (or boundaries $x$, $y$) of the image. If the chip shares one boundary with the image, we still check if the other side of the detection is completely enclosed inside or not. If it is not, we discard it, else we keep it. In another case, if the chip shares both the sides with the image boundary and so does the detection, then we keep the detection.

Once valid detections from each scale are obtained using the above rules, we merge detections from all the scales by projecting them to the image co-ordinates after applying appropriate scaling and translation. Finally, Non-Maximum Suppression is applied to aggregate the detections. The network architecture and an example of multi-scale inference and focus stacking is shown in Fig \ref{fig:framework}.

\subsection{Putting it all together}\label{ssec:together}
Here we summarize the proposed concepts and considerations leading up to the final scale-normalized object-detection paradigm to facilitate clear dissemination. First, we introduced the concept of scale-normalization for image-pyramids (SNIP) to effectively tackle the adverse effects of extreme-scale objects during training and put forward concrete guidelines for setting it's design parameters. This was followed by an efficient scale-normalized spatial sub-sampling mechanism (SNIPER) to reduce the additional computational cost involved in processing the multi-scale image-pyramid during training. We discussed the use of image content for SNIPER's spatial sub-sampling, described the positive/negative chip-sampling and their labeling scheme, and the additional benefits of SNIPER over SNIP with the help of increased batch-size and batch-normalization \cite{ioffe2015batch} for the training phase. Lastly, we modeled the active foveal-vision in humans in the form of AutoFocus that processes a multi-scale image-pyramid in a coarse-to-fine manner to reduce run-time computational cost during inference. We discussed the concept of FocusPixels that are used as a proxy for finding interesting regions which are the only spatial sub-regions where the detector needs to be applied during inference. This leads to the final system, which tackles scale-variation and can be efficiently trained and tested on multi-scale image-pyramids.
The proposed scale-normalization approach for object-detection makes several changes to the existing pipeline both during training and inference stages. Therefore, in this section, we employ the COCO dataset to carry out extensive ablation studies to clearly reveal the effects of different modules, namely SNIP, SNIPER, and AutoFocus which have been proposed. The comparisons with the other approaches follow standard protocol and use 123,000 images from the training and 20,288 images in test-dev set of COCO for training and evaluation. Since recall for proposals is not provided by the evaluation server on COCO, we train on 118,000 images and report recall on the remaining 5,000 images (commonly referred to as minival set, or the 2017 test-dev set). Unless specifically mentioned, the area of small objects is less than 32x32, medium objects range from 32x32 to 96x96 and large objects are greater than 96x96 pixels.
\section{Experimental Analysis}\label{sec:exp}

\subsection{Training Details}
\label{ssec:exp_train}
We use 3 resolutions of (480, 800), (800, 1200), and (1400, 2000) pixels for training our detectors. The first value is for the shorter side of the image and the second one is the limit on the maximum side size. The valid ranges are set to ($0$,$80^2$), ($32^2$, $150^2$), and ($120^2$, $\infty$) which ensure that at least 5 to 15 convolutional features are observed in the coarsest convolutional layer of the network. The training of the detectors is performed for 7 epochs. Except when RPN is trained separately, we use a shorter training period of 6 epochs. 

We start the training with a warmup learning rate of 0.0005 for 1000 iterations. Since we use the efficient resampling scheme, we can use a batch size of 128 chips for 512x512 pixels and a base learning rate of 0.015. When ablation experiments are performed for scale normalization, we use a batch size of 8 (1 per GPU) and a learning rate of 0.005. We use mixed-precision training as described in \cite{narang2017mixed}. To this end, we re-scale weight-decay by 100, drop the learning rate by 100, and re-scale gradients by 100. This ensures that we can train with activations of half-precision (and hence $\sim$2x larger batch size) without any drop of accuracy. FP32 weights are used for the first convolution layer, last convolution layer in RPN (classification and regression), and the fully connected layers in Faster-RCNN. We drop the learning rate after 5.33 epochs (except when RPN is trained separately where we drop the learning rate after 4.33 epochs). Image flipping is used as a data-augmentation technique.

As mentioned in Section \ref{sssec:sniper_neg}, an RPN is deployed for negative chip sampling in SNIPER. We train this RPN only for 2 epochs with a fixed learning rate of 0.015 without any step-down. Therefore, it requires less than 20\% of the total training time. RPN proposals are extracted from all scales. Note that inference takes 1/3 the time for a full forward-backward pass and we do not perform any flipping for extracting proposals. Hence, this process is also efficient.

\subsection{Effectiveness of SNIP against Scale Variation}\label{ssec:exp_snip_abl}
In this section, we carry different experiments to understand the behavior of SNIP under the variations of scale-range, object-detection architecture and to show the benefits of employing SNIP with other popular architectures. In order to disentangle and clearly understand the tolerance against scale-variation offered by SNIP in the two-stage object-detection pipelines, we evaluate the region-proposal and classification modules of the detection network separately.

\subsubsection{SNIP for RCN improvements}\label{sssec:exp_snip_abl_rcn}
In this ablation study, we use a single-scale proposal generation that is common across all the three scales of the multi-scale pyramid to generate the proposals. The generated proposals are used to evaluate the performance of SNIP on the RCN only, under the same settings as described in Section~\ref{subsec:snip}. This study aims at demonstrating the benefits of SNIP over the vanilla Multi-Scale Training/Testing pipeline. Therefore, we compare the performance of \emph{single-scale train/test},  \emph{multi-scale test} and \emph{multi-scale train/test} protocols against SNIP in Table \ref{tab:rcn_box}. For a fair comparison, we use the best possible validity ranges at each scale for all the protocols where multi-scale testing is performed. As expected, multi-scale testing yields an improvement of 1.4\% over single-scale train/test protocol. Naturally, we would expect that the multi-scale train/test protocol would improve it even further because multi-scale samples are used during training as well. However, it ended up reducing the improvement to only 1.1\% which clearly demonstrates that the inclusion of large objects (especially in the 1400$\times$2000 resolution) during multi-scale training adversely affects the training of the RCN. It happens due to the inability of the effective network receptive field to correctly classify extremely blown-up  objects in up-scaled images. Finally, SNIP improved the performance by 3.3\% and 1.9\% over single-scale train/test and multi-scale test protocols, respectively. This experiment clearly demonstrates the benefits of using SNIP during training to effectively avoid presenting large scale-variation to the RCN.

\begin{table}[t]
\begin{center}
\caption{MS denotes multi-scale. Single scale is (800,1200). R-FCN detector with ResNet-50 (as described in Section 4).}
\label{tab:rcn_box}
 \begin{tabular}{|c|c|c|c|c|c|c|c|c|c|}
  \hline
 \textbf{ Method} & \textbf{AP }&\textbf{ AP$^{S}$} & \textbf{AP$^{M}$} & \textbf{AP$^{L}$} \\
  \toprule
  Single scale & 34.5 & 16.3 & 37.2 & 47.6 \\
  MS Test & 35.9 & 19.5 & 37.3 & 48.5 \\
  MS Train/Test & 35.6 & 19.5 & 37.5 & 47.3 \\
  SNIP & 37.8 & 21.4 & 40.4 & 50.1 \\
  \bottomrule
 \end{tabular}
\end{center}

\end{table}

\begin{table}[t]
\begin{center}
 \caption{For individual ranges (like 0-25 etc.) recall at 50\% overlap is reported because minor localization errors can be fixed in the second stage. ResNet-50 is used as the backbone. Recall is for 900 proposals, as top 300 are taken from each scale.}
 \label{tab:rpn_box}
\begin{tabular}{|c|c|c|c|c|c|c|c|c|c|}
  \hline
  \textbf{Method} & \textbf{AR} & \textbf{AR$^{50}$} & \textbf{AR$^{75}$} & \textbf{0-25} & \textbf{25-50} & \textbf{50-100} \\
  \toprule
  Baseline & 61.3 & 89.2 & 69.8 & 68.1 & 91.0 & 96.7 \\
  + SNIP & 64.0 & 92.1 & 74.7 & 74.4 & 95.1 & 98.0 \\
  \bottomrule
 \end{tabular}
\end{center}

\end{table}

\subsubsection{SNIP for RPN improvements}\label{sssec:exp_snip_abl_rpn}
Now we turn our attention to the region-proposal network of the object detection pipeline. Before proceeding with the ablations studies with SNIP, note that the recall at 50\% overlap is the most important performance metric for object proposals; it's because bounding box regression can correct minor localization errors, but an uncovered object by all the proposals will certainly result in false negative. Since recall at 50\% overlap for the objects $>$100 pixel in size is already close to 100\%, further improvement wouldn't lead to any significant overall improvements. Improving the recall on small objects, however, would lead to more overall gains. In order to demonstrate the benefits of SNIP training on differently sized objects, we show the improvements for a ResNet-50 RPN network in Table \ref{tab:rpn_box}. First, note that SNIP improved the overall recall at 50\% overlap by 2.9\% and 6.3\% for objects smaller than 25 pixels. If we train our RPN without SNIP, mAP drops by 1.1\% on small objects and 0.5\% overall. Note that AP of large objects is not affected as we still use the classification model trained with SNIP. We also perform an ablation study with stronger backbones like DPN-98 and detectors like Faster-RCNN which are shown in Table \ref{tab:snip_ablations}.


\begin{table}[t]
\begin{centering}
 \caption{The effect of SNIP on RCN and RPN}
 \label{tab:snip_ablations}
\begin{tabular}{|c|c|c|c|c|}
  \hline
  \textbf{Method} & \textbf{Backbone} & \textbf{RPN SNIP} & \textbf{RCN SNIP} & \textbf{AP} \\
  \toprule
  \multirow{3}{*}{D-R-FCN} & \multirow{3}{*}{DPN-98}  & \xmark & \xmark & 41.2\\
    & & \xmark & \cmark & 44.2  \\
    & & \cmark & \cmark & 44.7  \\
  \midrule
  \multirow{3}{*}{Faster-RCNN} & \multirow{3}{*}{ResNet-101} & \xmark & \xmark & 42.6 \\
    & & \cmark & \xmark & 43.1  \\
    & & \cmark & \cmark & 44.4  \\
  \bottomrule
 \end{tabular}

 \end{centering}
\end{table}

\subsection{Efficient Training with SNIPER}\label{ssec:exp_sniper_abl}
While SNIP improves training on multi-scale image pyramid, it comes at a computational cost. In this section, we carry out ablation studies pertaining to the efficient re-sampling scheme for training with SNIP (or SNIPER). As SNIPER does not use all the training data and uses RPN to generate chips for training, we have to ensure that RPN recall is good. Moreover, we lose a significant amount of background samples during training, therefore, it's important to assess the effect of negative chip mining. The final goal of design parameters is to ensure that SNIPER's performance matches the SNIP baseline, which trains on the entire image pyramid, while obtaining a significant speedup. In the following sub-sections, we focus on different analyses pertaining to the design parameters of SNIPER to achieve the aforementioned goal.

\subsubsection{SNIPER Recall Analysis}\label{sssec:exp_sniper_abl_recall}
Since the positive chip-sampling covers all the ground truth samples, we posit that it's sufficient to train on just the positive samples for generating proposals and still maintain a high recall. Indeed, we observe that the recall (averaged over multiple overlap thresholds 0.5:0.05:0.95) for RPN is unaffected w.r.t. negative sampling (Table~\ref{tab:sniper_negmining_recall}) because recall doesn't account for false positives. The aforementioned intuitive reasoning and empirical results bolster SNIPER's strategy of employing an RPN, which is trained on positive samples only, for negative chip-sampling. However, mAP score for detection depends on false-positives, as shown in Sec.~\ref{sssec:sniper_neg}, hence negative sampling, discussed next, plays an important role as well.

\begin{table}[t]
\begin{center}
 \caption{We plot the recall for SNIPER with and without negatives. Surprisingly, recall is not effected by negative chip sampling}
\label{tab:sniper_negmining_recall}
\begin{tabular}{|c|c|c|c|c|c|c|c|c|c|c|}
  \hline
  \textbf{NEG.} & \textbf{AR} & \textbf{AR$^{50}$} & \textbf{AR$^{75}$} & \textbf{0-25} & \textbf{25-50} & \textbf{50-100}  & \textbf{100-300}\\
  \toprule
  \cmark & 65.4 & 93.2 & 76.9 & 41.3 & 65.8 & 74.5 & 77.8 \\
  \xmark & 65.4 & 93.2 & 77.6 & 40.8 & 65.7 & 74.7 & 78.3 \\
  \bottomrule
 \end{tabular}
\end{center}
\end{table}

\begin{table}[t]
\begin{center}
\caption{The effect training on 2 scales (1.667 and max size of 512). We also show the impact in performance when no negative mining is performed. A ResNet-101 backbone is used.}
\label{tab:neg_mining}
\begin{tabular}{|c|c|c|c|c|c|c|c|c|}
  \hline
  \textbf{Method} & \textbf{AP} & \textbf{AP$^{50}$} & \textbf{AP$^{75}$} & \textbf{AP$^{S}$} & \textbf{AP$^{M}$} & \textbf{AP$^{L}$} \\
  \toprule
  SNIPER   & 46.1 & 67.0 & 51.6 & 29.6 & 48.9 & 58.1 \\
  No Neg.  & 43.4 & 62.8 & 48.8 & 27.4  & 45.2 & 56.2 \\
  2 Scales    & 43.3 & 63.7 & 48.6 & 27.1  & 44.7 & 56.1 \\
  \bottomrule   
 \end{tabular}
\end{center}
\end{table}

\begin{table}[t]
\begin{center}
\caption{We observe that SNIPER matches the performance even after reducing the pixels processed by 3.5 times.}
\label{tab:snip_comp}
\begin{tabular}{|c|c|c|c|c|c|c|c|c|}
  \hline
  \textbf{Method} & \textbf{AP} & \textbf{AP$^{50}$} & \textbf{AP$^{75}$} & \textbf{AP$^{S}$} & \textbf{AP$^{M}$} & \textbf{AP$^{L}$} \\
  \toprule
  SNIP   & 43.6 & 65.2 & 48.8 & 26.4 & 46.5 & 55.8 \\
  SNIPER  & 43.5 & 65.0 & 48.6 & 26.1 & 46.3 & 56.0 \\
  \bottomrule   
 \end{tabular}
\end{center}

\end{table}

\begin{table}[t]
\begin{center}
\caption{We highlight the importance of image pyramids even with lightweight backbones, where we see a 12\% gain in performance. Pre-training with additional data and multi-tasking with instance segmentation brings a 1.5\% improvement in performance.}
\label{tab:sniper_comp}
\begin{tabular}{|c|c|c|}
  \hline
  \textbf{Method} & \textbf{Backbone} & \textbf{AP}  \\
  \toprule
  SSD   & MobileNet-v2 & 22.1 \\
  SNIPER   & MobileNet-v2 & 34.5 \\
  \midrule
  SNIPER & ResNet-101 & 46.1 \\
  SNIPER + OpenImages & ResNet-101 & 46.8 \\
  SNIPER + OpenImages + Mask Training & ResNet-101 & 47.6 \\
  \bottomrule   
 \end{tabular}
\end{center}

\end{table}

\subsubsection{Effect of Negative Chip Mining on SNIPER}
\label{sssec:exp_sniper_abl_neg}
Just like any other object-detection system, SNIPER also employs negative chip mining to reduce the false-positive rate. Additionally, SNIPER also aims at speeding-up the training by skipping the \textit{easy} regions inside the image, which are obtained from an RPN trained with a short learning schedule, Sec.~\ref{sssec:sniper_neg}. Inclusion of negative samples that are similar in appearance to positive instances is a well-known technique to reduce the false-positive rate and helps to improve the overall mAP, which depends on both the recall and precision. To evaluate the effectiveness of our negative mining approach, we compare SNIPER's mAP score with a variant that only uses positive chips during training, Table~\ref{tab:neg_mining}, while keeping other parameters the same. The proposed negative chip mining approach noticeably improves AP scores for all localization thresholds and object sizes and improves the mAP score from $43.4$ to $46.1$.

\subsubsection{Effect of Multi-Scale Training on SNIPER}
In order to illustrate the benefits of multi-scale training using SNIPER, we reduce the number of scales from $3$ to $2$ by dropping the highest resolution scale and trained with SNIPER. This variant is compared with  standard SNIPER training that employs all $3$ scales for training and the results are compared in Table~\ref{tab:neg_mining}. We can see that the reduction in the number of scales significantly decreased the performance consistently across all evaluation metrics. 

\subsubsection{Comparison with training on Full Image Pyramids}\label{sssec:exp_sniper_abl_time}
Since SNIPER reduces both the memory and computational footprint while processing a multi-scale image pyramid, it affords increased batch-size and effective batch-normalization \cite{ioffe2015batch} during training, which was otherwise not possible with SNIP on commodity GPU cards. Therefore, in order to compare SNIPER with SNIP, we turn-off batch-normalization during SNIPER training and show that it achieves matching results with SNIP, in Table~\ref{tab:snip_comp}. With batch-normalization, SNIPER significantly outperforms SNIP in all metrics and obtains an mAP of 46.1\%. This result improves to 46.8\% if we pre-train the detector on the OpenImagesV4 dataset. Adding an instance segmentation head and training the detection network along with it further improves the performance to 47.6\%. We also show results for Faster-RCNN trained with MobileNetV2. It obtains an mAP of 34.1\% compared to the SSDLite \cite{mobilenetv2} version which obtained 22.1\%. This again highlights the importance of image pyramids (and SNIPER training) as we can improve the performance of the detector by 12\%, Table~\ref{tab:sniper_comp}. Not only is SNIPER more accurate, it is also $3\times$ faster compared to SNIP during training. It only takes 14 hours for end-to-end training on a 8x V100 GPU node with a Faster-RCNN detector with ResNet-101 backbone. It is worth noting that we train on 3 scales of an image pyramid (max size of 512, 1.667 and 3). Training RPN is much more efficient and it only takes 2 hours.

\begin{figure}
\centering
\includegraphics[width=0.6\linewidth]{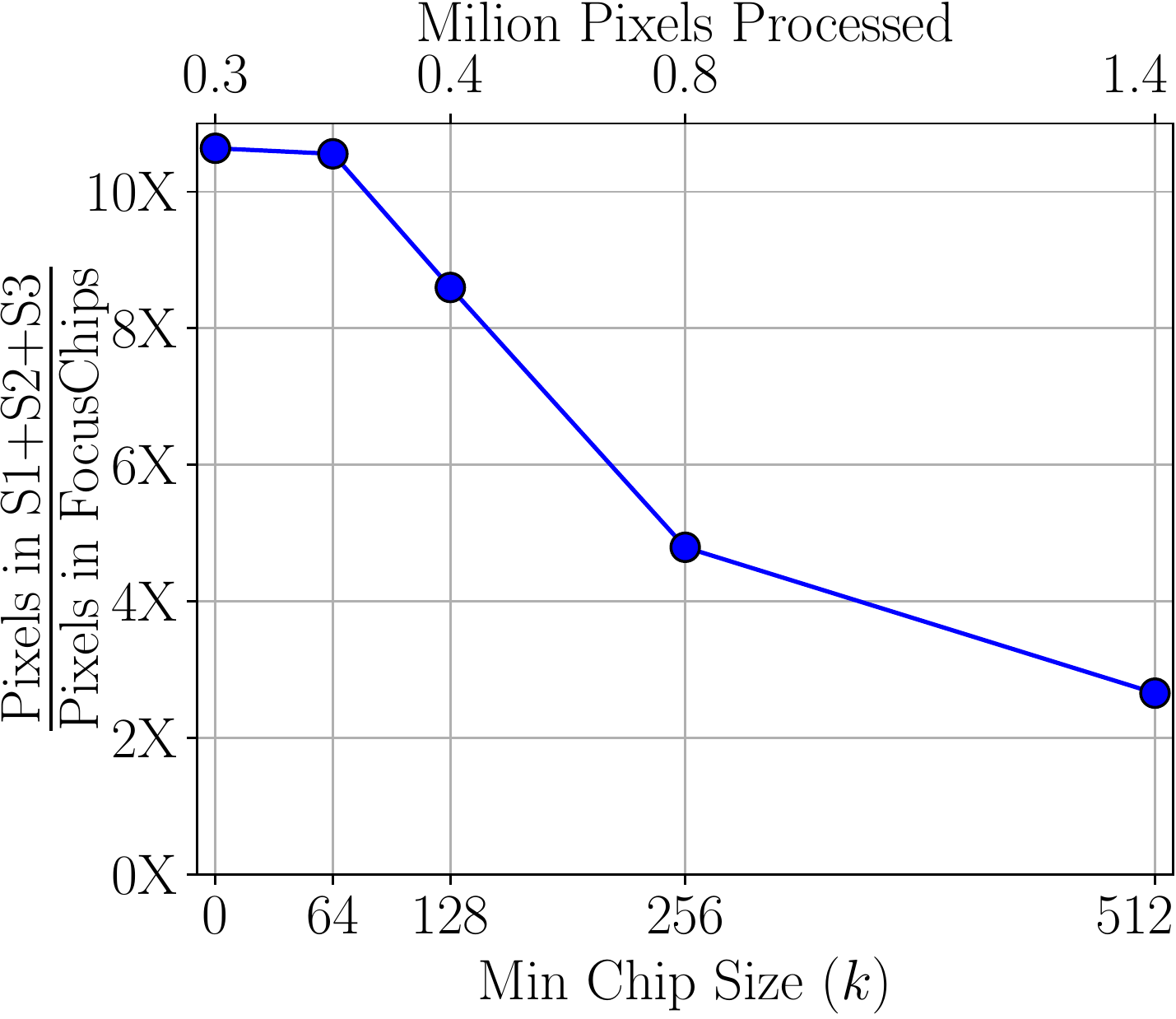}
\caption{Upper-bound on the speed-up using FocusChips generated from optimal FocusPixels.}
\label{fig:upperbound}
\end{figure}

\begin{figure*}
    \centering
    \begin{subfigure}[t]{0.261\linewidth}
    \includegraphics[width=\linewidth]{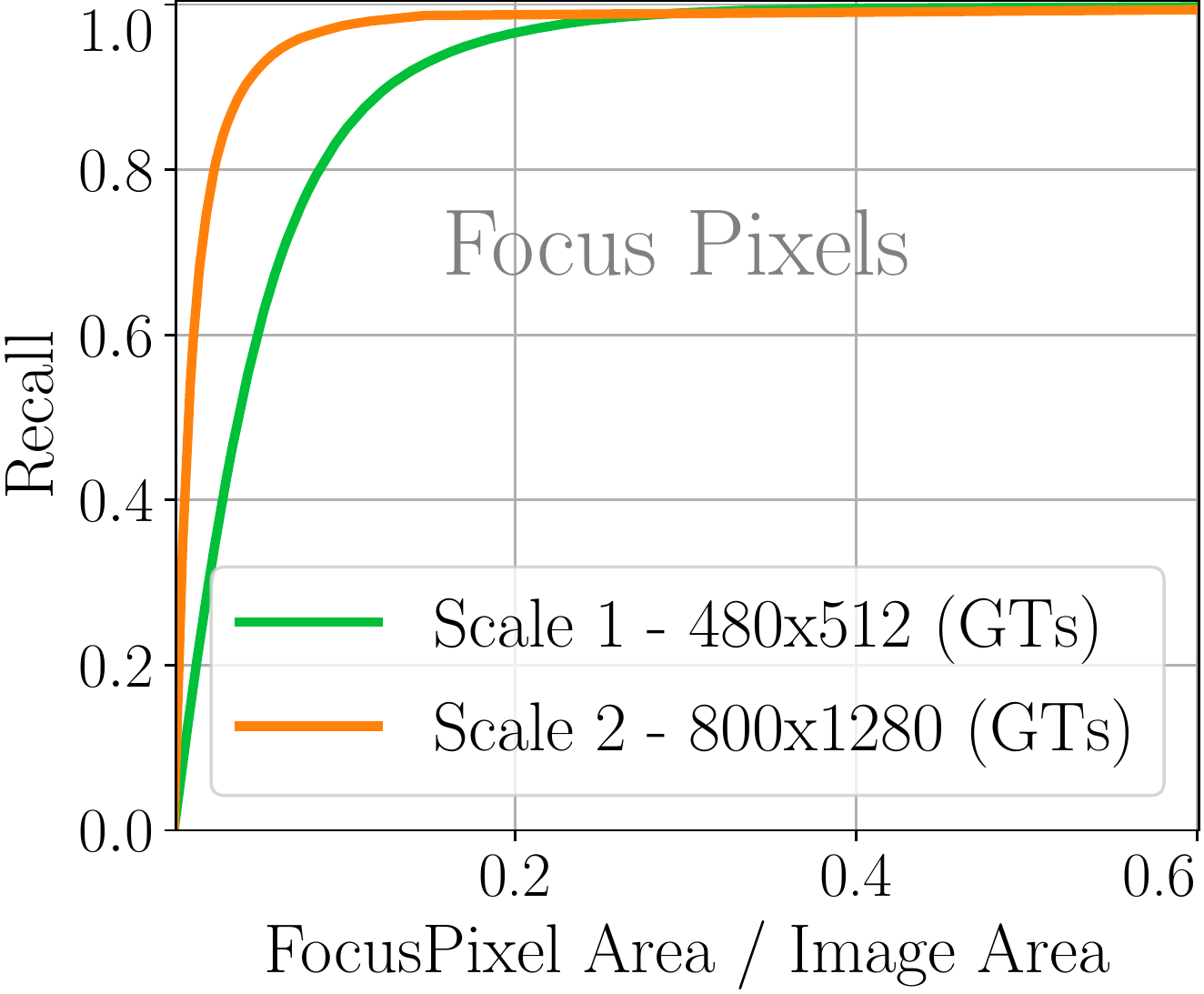}
    \caption{}
    \end{subfigure}
    \begin{subfigure}[t]{0.241\linewidth}
    \includegraphics[width=\linewidth]{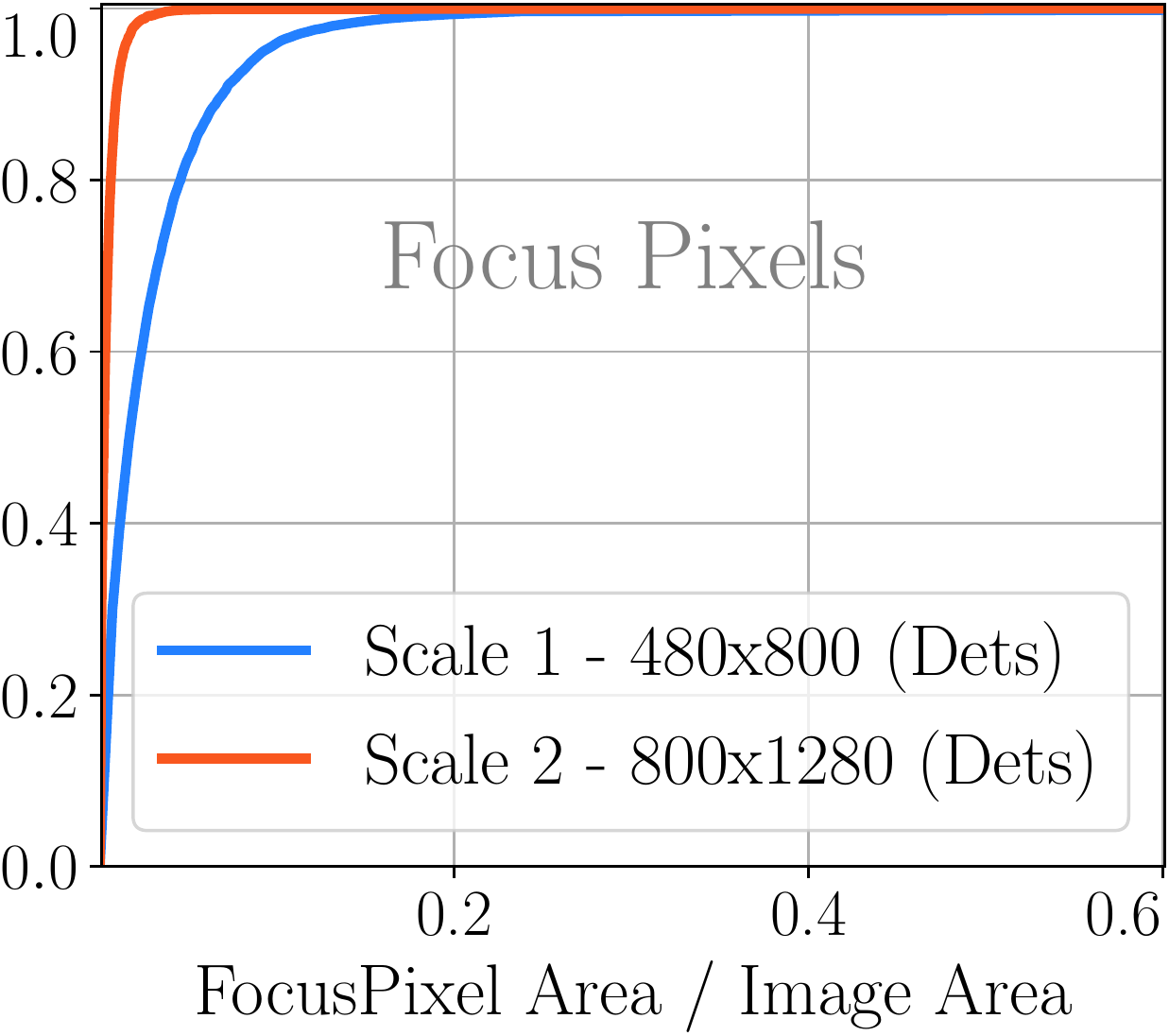}
    \caption{}
    \end{subfigure}
      \begin{subfigure}[t]{0.241\linewidth}
    \includegraphics[width=\linewidth]{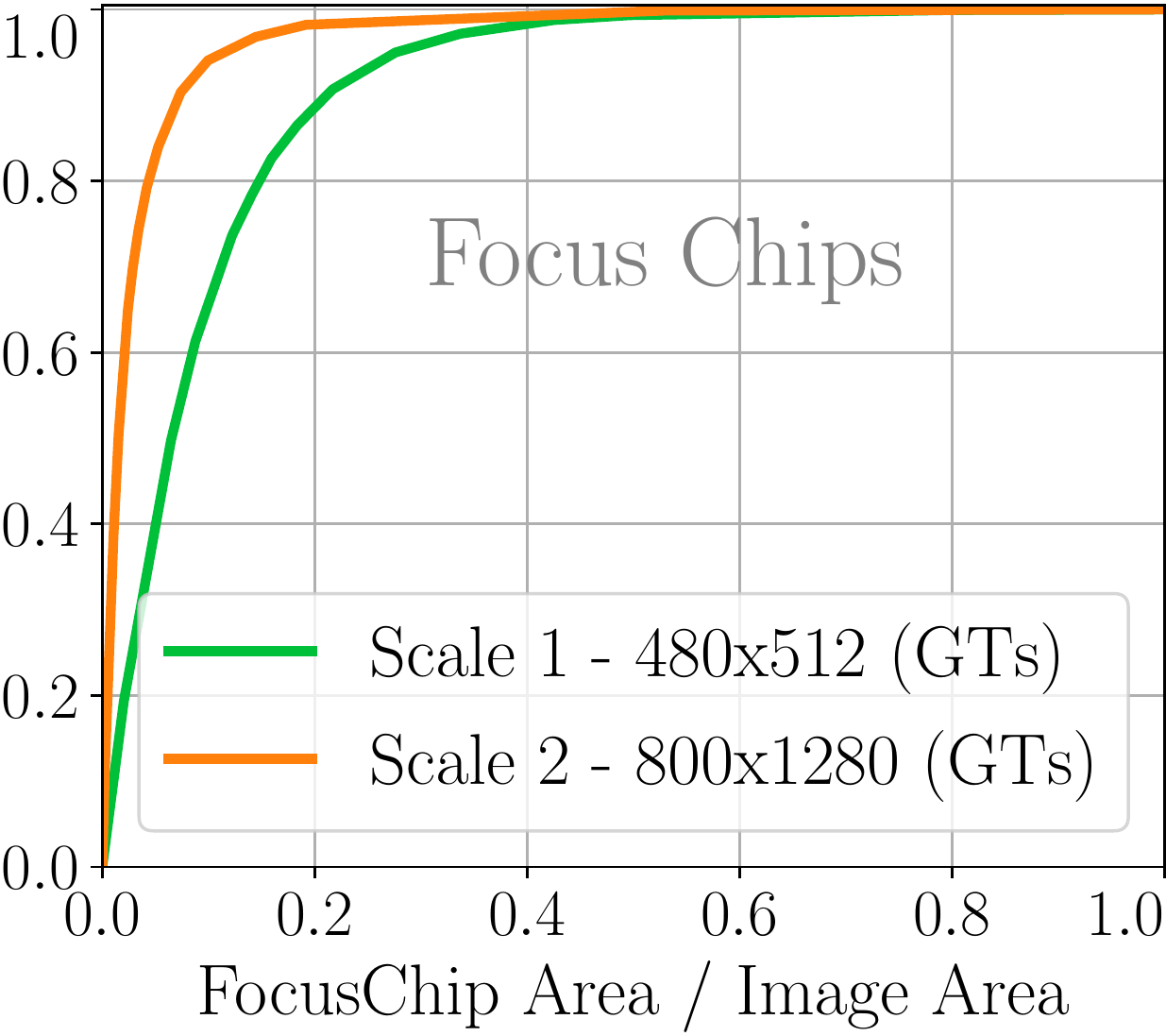}
    \caption{}
    \end{subfigure}
    \begin{subfigure}[t]{0.241\linewidth}
    \includegraphics[width=\linewidth]{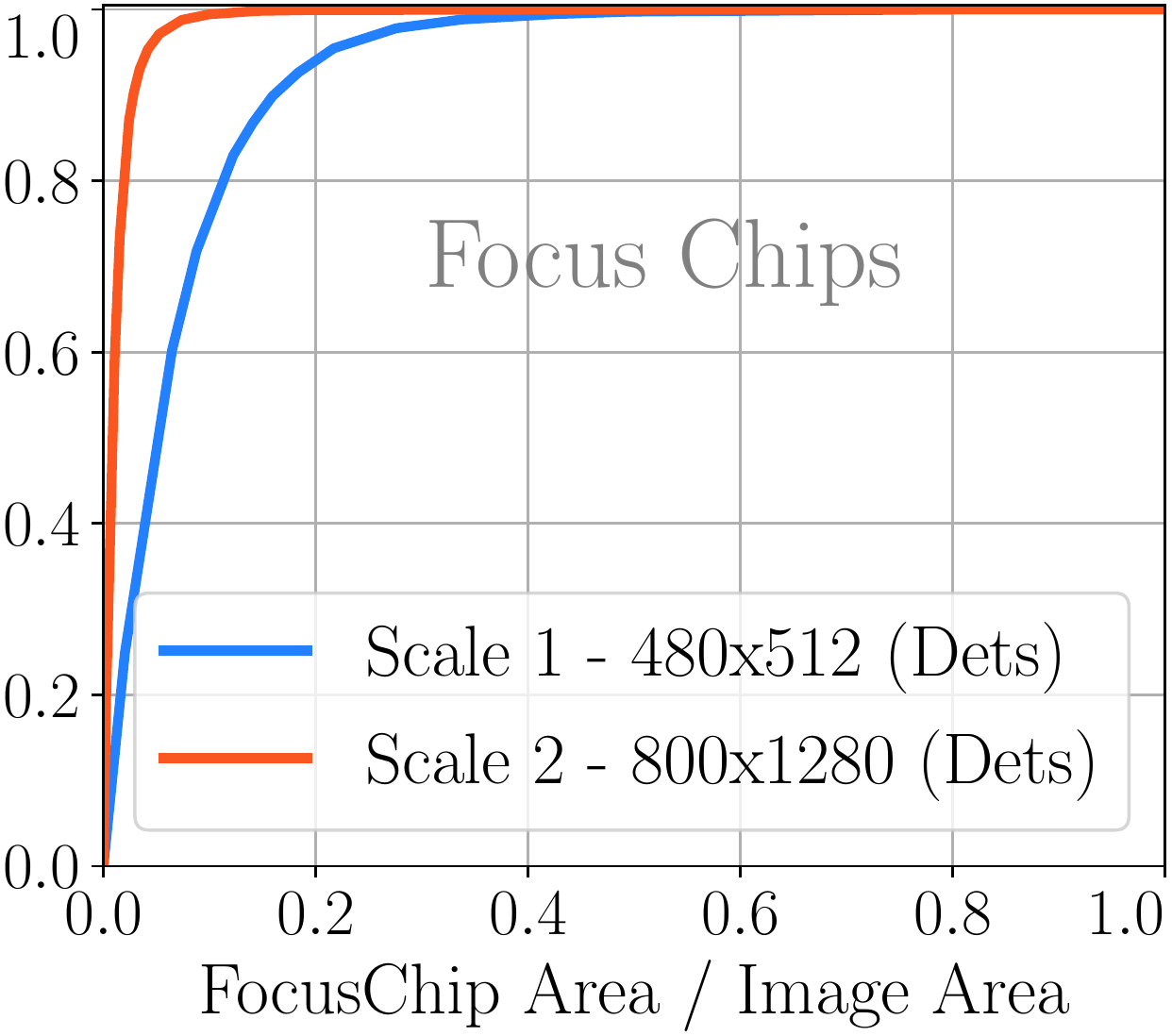}
    \caption{}
    \end{subfigure}
    \caption{Quality of the FocusPixels and FocusChips. The x-axis represents the ratio of the area of FocusPixels or FocusChips to that of the image. The y-axis changes as follows, (a) FocusPixel recall is computed based on the GT boxes (b) FocusPixel recall is computed using the confident detections (c) FocusChip recall is computed based on the GT boxes (d) FocusChip recall is computed based on the confident detections.}
    \label{fig:auto_curves}
\end{figure*}

\subsection{Efficient Inference with AutoFocus}\label{ssec:exp_auto_abl}
While SNIPER improves the efficiency of training by skipping ``easy'' regions, it is not directly applicable during inference as it requires ground-truth information for chip sampling. As discussed in \ref{subsec:auto}, AutoFocus extends the active sub-sampling concept to the inference phase by predicting ``FocusPixels'' and generating ``FocusChips'' from them. In this section, we empirically -- evaluate the underlying hypotheses behind AutoFocus and the quality of the estimated FocusPixels/Chips w.r.t. the ground-truth, compare AutoFocus with SNIPER, and study the Speed-Accuracy trade-off w.r.t. design parameters.

\subsubsection{AutoFocus Hypothesis Testing}\label{sssec:exp_auto_abl_stat}
The core hypothesis behind AutoFocus is the low percentage of the FocusPixels in natural images, especially in high-resolution images. To investigate this hypothesis, here, we report the percentage of the FocusPixels at different scales for the validation set of the COCO dataset based on ground-truth annotations. In high-resolution images (scale 3), the percentage of FocusPixels is very low (\ie $\sim4$\%). Therefore, ideally, a very small part of the image needs to be processed at high resolution. Since the image is up-sampled, the FocusPixels projected on the image occupy an area of $63^2$ pixels on average (the highest resolution images have an area of $1602^2$ pixels on average). At lower scales (like scale 2), although the percentage of FocusPixels increases to $\sim11\%$, their projections only occupy an area of $102^2$ pixels on average (each image at this scale has an average area of $940^2$ pixels). After dilating FocusPixels with a kernel of size 3 $\times$ 3, their percentages at scale 3 and scale 2 change to 7\% and 18\% respectively.

Using the chip generation algorithm, for a given minimum chip size (like $k=512$), a theoretical upper bound on the speedup can be obtained under the assumption that FocusPixels can be predicted without any error (\ie based on GTs). This speedup bound changes with the minimum chip size and this variation is shown in Fig \ref{fig:upperbound}, following the FocusChip generation algorithm \ref{alg:chip_generation}. The same value is used at each scale. For example, reducing the minimum chip size from 512 to 64 can lead to a theoretical speedup of $\sim10$ times over the baseline which performs inference on 3 scales. However, a significant reduction in minimum chip size can also affect detection performance - a reasonable amount of context is necessary for retaining high detection accuracy.

\subsubsection{Quality of FocusPixel prediction}
\label{sssec:focus_pixel_quality}
Here, we evaluate how well our network predicts FocusPixels at different scales using two criteria. First, we measure the recall for predicting FocusPixels at two different resolutions and show the results in Fig \ref{fig:auto_curves} \textcolor{red}{a}. This provides us with an upper bound on the accuracy of localizing small objects using low-resolution images. However, not all the ground-truth objects that are annotated might be correctly detected. Since our eventual goal is to accelerate the detector, cropping regions that cover ground-truth instances which the detector cannot detect would not be useful. Therefore, the final effectiveness of FocusChips is \emph{intrinsically} coupled with the detector, hence we also report the accuracy of FocusPixel prediction on regions which are confidently detected in Fig \ref{fig:auto_curves} \textcolor{red}{b}. This is achieved by only considering the FocusPixels corresponding to the GT boxes that significantly overlap (IoU $>$ 0.5) with a detection-box with a score $\le 0.5$. At a threshold of 0.5, the detector still obtains an mAP of 47\% which is within 1\% of the final mAP and does not have a high false-positive rate.

As expected, we obtain better recall at higher resolutions with both metrics. We can cover all confident detections at the higher resolution (scale 2) when the predicted FocusPixels cover just 5\% of the total image area. At a lower resolution (scale 1), when the FocusPixels cover 25\% of the total image area, we cover all confident detections, see Fig \ref{fig:auto_curves} \textcolor{red}{b}. 

\subsubsection{Quality of FocusChips}\label{sssec:exp_auto_abl_qual}
Eventually, it's the FocusChips and not FocusPixels that are input to the network, therefore, we evaluate the accuracy of the generated FocusChips, from the FocusPixels, using similar metrics as in Sec.~\ref{sssec:focus_pixel_quality} - the recall of all GT boxes enclosed by FocusChips and the recall for GT boxes enclosed by FocusChips that overlap with a confident detection. To achieve perfect recall for confident detections at scale 2, FocusChips cover 5\% more area than FocusPixels. At scale 1, they cover 10\% more area. This is because objects are often not rectangular in shape. These results are shown in Fig \ref{fig:auto_curves} \textcolor{red}{d}.

\begin{figure*}
    \centering
    \begin{subfigure}[t]{0.245\linewidth}
    \includegraphics[width=\linewidth]{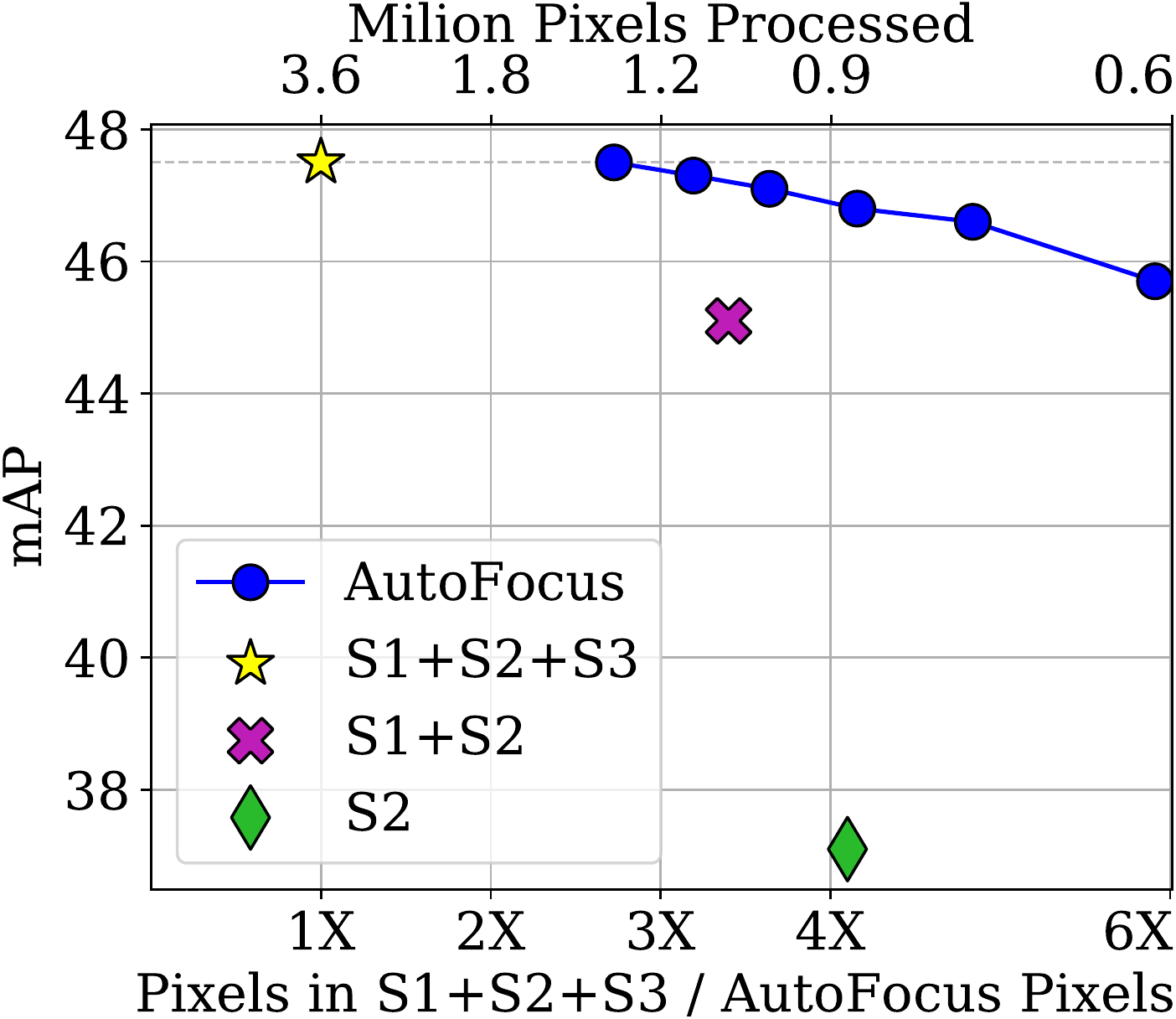}
    \caption{}
    \end{subfigure}
    \begin{subfigure}[t]{0.245\linewidth}
    \includegraphics[width=\linewidth]{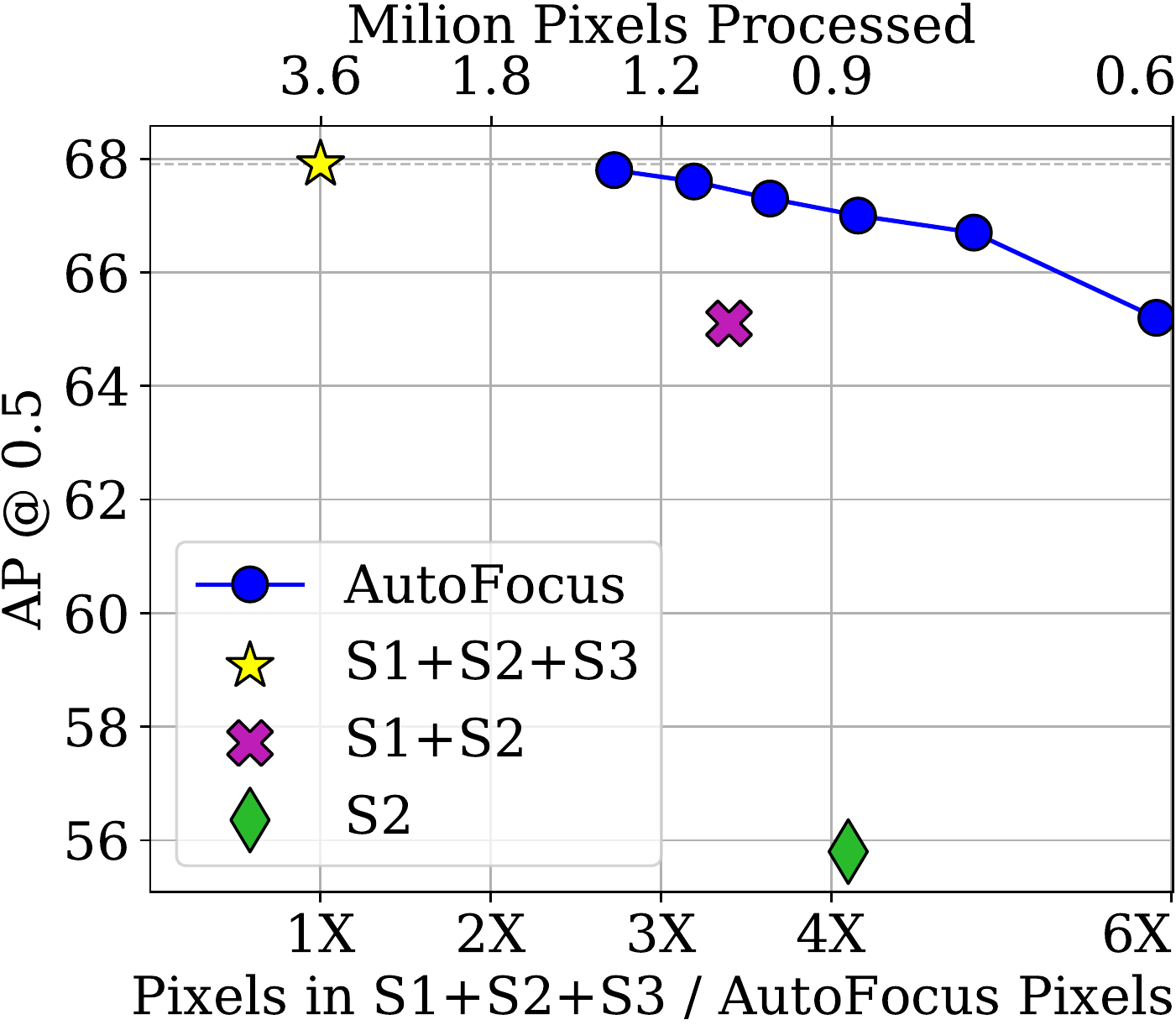}
    \caption{}
    \end{subfigure}
      \begin{subfigure}[t]{0.25\linewidth}
    \includegraphics[width=\linewidth]{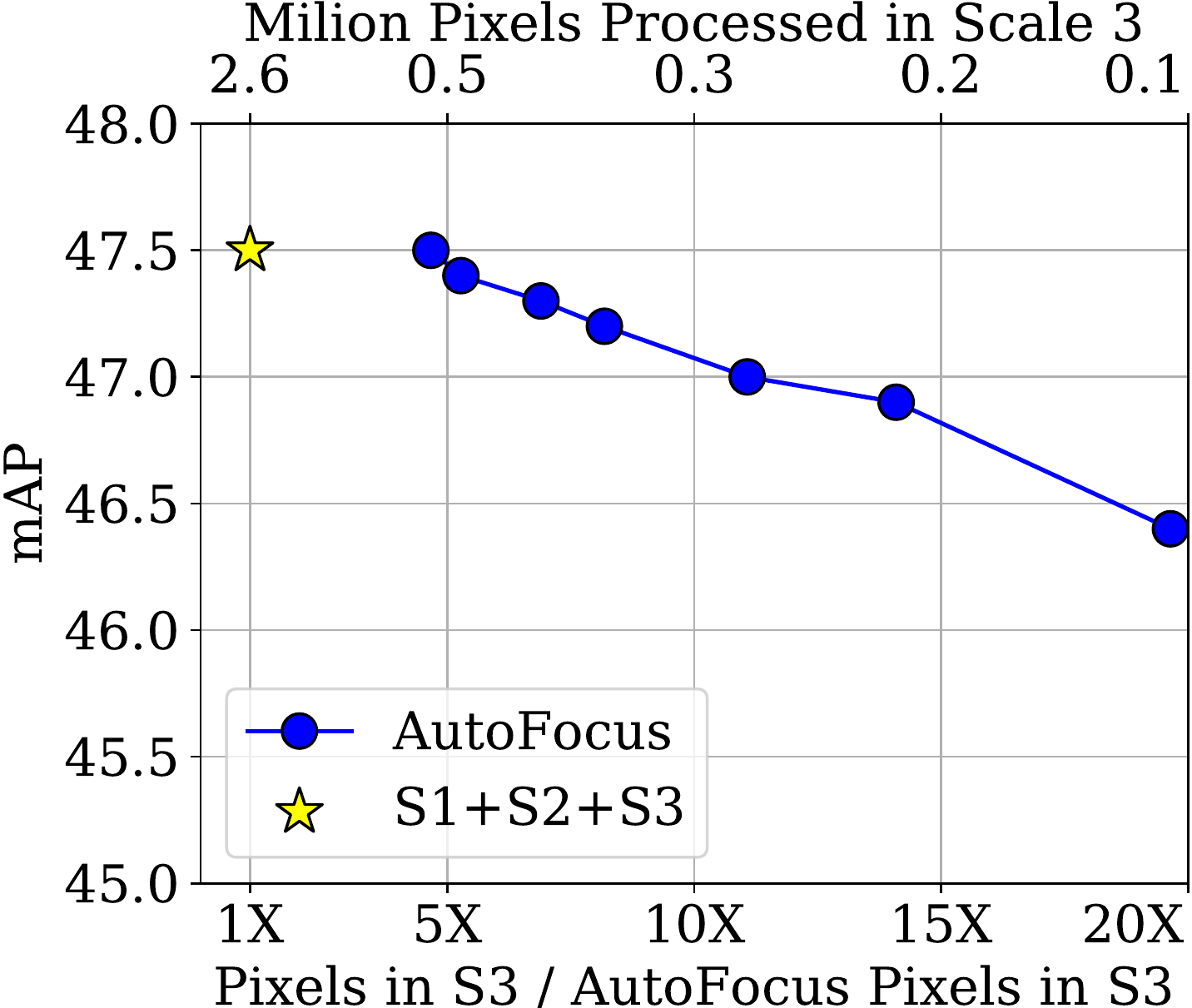}
    \caption{}
    \end{subfigure}
    \begin{subfigure}[t]{0.245\linewidth}
    \includegraphics[width=\linewidth]{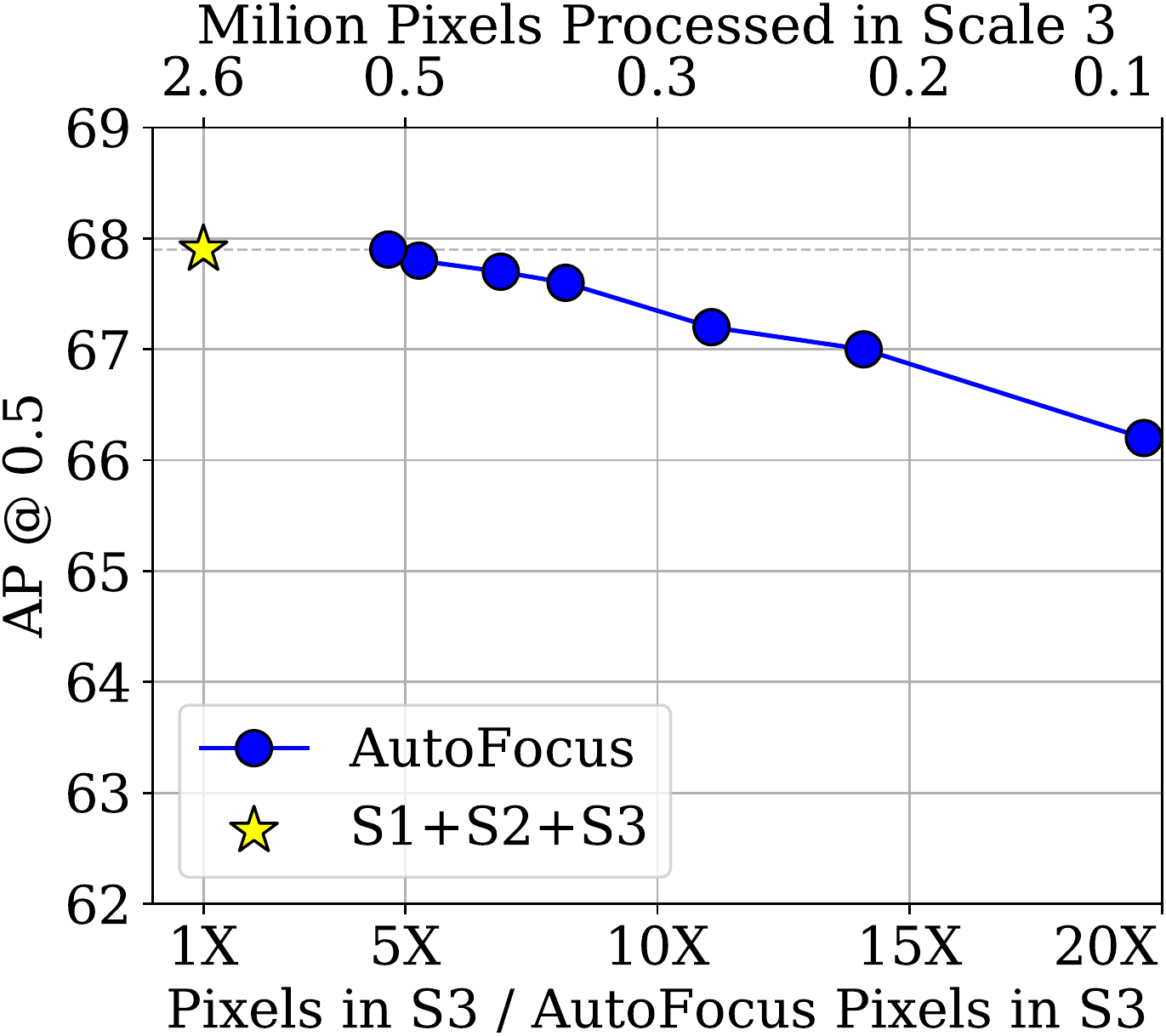}
    \caption{}
    \end{subfigure}
    \caption{Results are on the val-2017 set. (a,c) show the mAP averaged for IoU from 0.5 to 0.95 with an interval of 0.05 (COCO metric). (b,d) show mAP at 50\% overlap (PASCAL metric). We can reduce the number of pixels processed by a factor of 2.8 times without any loss of performance. A 5 times reduction in pixels is obtained with a drop of 1\% in mAP. }
    \label{fig:speed_acc}
\end{figure*}

\subsubsection{Comparison with SNIPER}\label{sssec:exp_auto_sniper_comparison}
In this section, we compare the efficient multi-scale inference in AutoFocus, with testing on Full Image Pyramids in SNIPER. For fair comparison, in AutoFocus, we just add the FocusPixel prediction branch to our detector while keeping everything else the same. Table \ref{tab:auto_sniper_comparison} shows the results.  While matching SNIPER's performance of 47.9\% (68.3\% at 0.5 IoU), AutoFocus processes 6.4 images per second on the test-dev set with a TitanX Pascal GPU. SNIPER processes 2.5 images per second. Moreover, AutoFocus is able to reduce the average number of pixels processed to half while dropping the AP by just $0.7$\%. This shows the effectiveness of the chip sampling process in AutoFocus.

\begin{table}[t]
\begin{center}
\caption{Comparison between the efficient multi-scale testing in AutoFocus and testing on full image pyramids in SNIPER on COCO test-dev. The average pixels processed over the  dataset  are  also reported.}
\label{tab:auto_sniper_comparison}
\begin{tabular}{|c|c|c|c|c|c|c|c|c|}
  \hline
  \textbf{Method} & \textbf{Pixels} & \textbf{AP} & \textbf{AP}$^\textbf{50}$ & \textbf{S} & \textbf{M} & \textbf{L} \\
  \toprule
   SNIPER  & 1910$^2$ & \textbf{47.9} & \textbf{68.3} & \textbf{31.5} & \textbf{50.5} & \textbf{60.3} \\ 
 \midrule
   \multirow{2}{*}{AutoFocus} & 1175$^2$ & \textbf{47.9} & \textbf{68.3} & \textbf{31.5} & \textbf{50.5} &  \textbf{60.3} \\  
  & 930$^2$ & 47.2 & 67.5 & 30.9 & 49.0 & 60.0 \\
  \bottomrule   
 \end{tabular}
\end{center}

\end{table}

\subsubsection{Speed Accuracy Trade-off for AutoFocus}\label{sssec:exp_auto_abl_trade}
In Section \ref{sssec:exp_auto_sniper_comparison}, we showed that AutoFocus inference accuracy matches that of the SNIPER's. Moreover, as discussed in Section \ref{subsec:auto}, the speed accuracy trade-off in AutoFocus can be further controlled. Therefore, we study the effect of AutoFocus parameters on its inference speed and accuracy. We perform a grid-search on the concerned parameters - dilation, min-chip size and the threshold - to generate FocusChips on a subset of 100 images in the validation set. For a given average number of pixels, we check which configuration of parameters obtains the best mAP on this subset. Since there are two scales at which we predict FocusPixels, we first find the parameters of AutoFocus when it is only applied to the highest resolution scale. Then we fix these parameters for the highest scale, and find parameters for applying AutoFocus at scale 2. 

In Fig \ref{fig:speed_acc} we show that the multi-scale inference baseline which uses 3 scales obtains an mAP of 47.5\% (and 68\% at 50\% overlap) on the val-2017 set. Using only the lower two scales obtains an mAP of 45.4\%. The middle scale alone obtains an mAP of 37\%. This is partly because the detector is trained with our scale normalization scheme. As a result, the performance on a single scale alone is not very good, although multi-scale performance is high. The maximum savings in pixels which we can obtain while retaining performance is 2.8 times. We lose approximately 1\% mAP to obtain a 5 times reduction over our baseline in the val-2017 set.

We also perform an ablation experiment for the FocusPixels predicted using scale 2. Note that the performance of just using scales 1 and 2 is 45\%. We can retain the original performance of 47.5\% on the val-2017 set by processing just one-fifth of scale 3. With a 0.5\% drop, we can reduce the pixels processed by 11 times in the highest resolution image. This can be improved to 20 times with a 1\% drop in mAP, which is still 1.5\% better than the performance of the lower two scales.

\begin{figure*}[t]
    \centering
    \includegraphics[width=\linewidth]{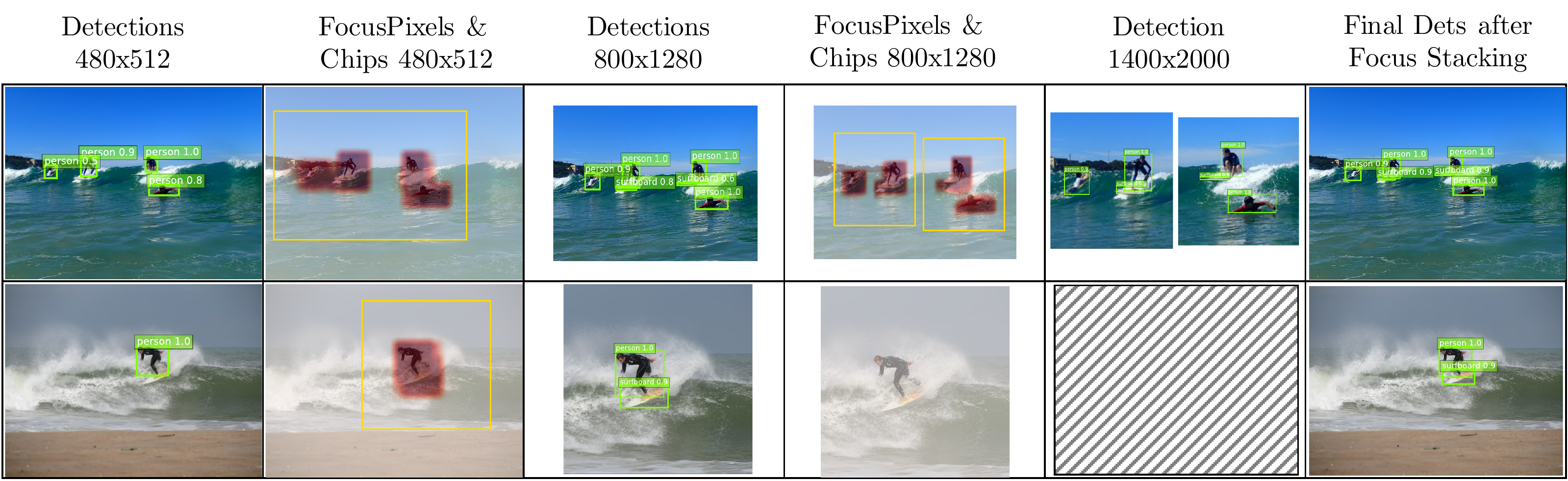}
    \caption{Each row shows the inference pipeline in AutoFocus. The confidence for FocusPixels and FocusChips are shown in red, and yellow respectively in the second and fourth columns. Detections are shown in green. As can be seen, complex images containing many small objects like the first row can generate multiple FocusChips in high resolutions like $1400\times2000$. Images which do not contain small objects are not processed at all in high resolution, like the one in the second row.}
    \label{fig:quals}
\end{figure*}

\begin{table}[!ht]
\centering
\caption{Comparison on the COCO test-dev. Results for others are taken from the papers/GitHub of the authors. Note that average pixels processed over the dataset are reported (instead of the shorter side). All methods use a ResNet-101 backbone. `+' denotes the multi-scale version provided by the authors.}
 \label{tab:final_coco}

\begin{tabular}{|c|c|c|c|c|c|c|}
\hline
  \textbf{Method} & \textbf{Pixels} & \textbf{AP} & \textbf{AP}$^\textbf{50}$ & \textbf{S} & \textbf{M} & \textbf{L} \\
  \toprule
  Retina \cite{lin2017focal} & 950$^2$ & 37.8 & 57.5 & 20.2 & 41.1 & 49.2 \\
  D-RFCN \cite{bodla2017} & 950$^2$ & 38.4 & 60.1 &18.5 & 41.6 & 52.5 \\
  Mask-RCNN \cite{he2017mask} & 950$^2$ & 39.8 & 62.3 & 22.1 & 43.2 & 51.2\\
  FSAF\cite{zhu2019feature} & 950$^2$ & 40.9 & 61.5 & 24.0 & 44.2 & 51.3\\
  LightH \cite{li2017light} & 950$^2$ & 41.5 & - & 25.2 & 45.3 & 53.1 \\
  FCOS\cite{tian2019fcos} & 950$^2$ & 41.5 & 60.7 & 24.4 & 44.8 & 51.6\\
  Refine+ \cite{zhangsingle} & 3100$^2$ & 41.8 & 62.9 & 25.6 & 45.1 & 54.1 \\
  Corner+ \cite{law2018cornernet} & 1240$^2$ & 42.1 & 57.8 & 20.8 & 44.8 & 56.7 \\
  FoveaBox-align \cite{kong2019foveabox} & 950$^2$ & 42.1 & 62.7 & 25.2 & 46.6 & 54.5 \\
  Cascade R-CNN\cite{cai2017cascade}&  950$^2$& 42.8& 62.1& 23.7& 45.5& 55.2\\
  FSAF+ (+hflip) \cite{zhu2019feature} & 4100$^2$ & 42.8 & 63.1 & 27.8 & 45.5  & 53.2\\
  RepPoints\cite{yang2019reppoints} & 950$^2$& 45.0& 66.1&  26.6 & 48.6 & 57.5\\ 
  RepPoints+\cite{yang2019reppoints} & 2850$^2$ & 46.5 & 67.4 &  30.3 & 49.7  & 57.1\\
  
  \midrule
  SNIP & 1910$^2$ & 44.4 & 66.2 & 27.3 & 47.4 & 56.9\\
  \midrule
 SNIPER  & 1910$^2$ & \textbf{47.9} & \textbf{68.3} & \textbf{31.5} & \textbf{50.5} & \textbf{60.3} \\ 
 \midrule
    & 1175$^2$ & \textbf{47.9} & \textbf{68.3} & \textbf{31.5} & \textbf{50.5} &  \textbf{60.3} \\  
  AutoFocus   & 930$^2$ & 47.2 & 67.5 & 30.9 & 49.0 & 60.0 \\
     & 860$^2$ & 46.9 & 67.0 & 30.1 & 48.9 & 60.0 \\
 \bottomrule
 \end{tabular}

\end{table}

\begin{table}[!ht]
\caption{Comparison on the PASCAL VOC 2007 test-set. All methods use ResNet-101 and trained on VOC2012 trainval+VOC2007 trainval. The average pixels processed over the  dataset  are  also reported. To show the robustness of AutoFocus to hyper-parameter choices, in `*' we use the same parameters as COCO and run the algorithm on PASCAL.}
 \label{tab:final_voc}
\centering
\begin{tabular}{|c|c|c|c|}
\hline
  \textbf{Method} & \textbf{Pixels} & \textbf{AP}$^\textbf{50}$ & \textbf{AP}$^\textbf{70}$ \\
  \toprule
  Faster RCNN\cite{ren2015faster} & 705$^2$& 76.4 & - \\
  R-FCN \cite{dai2016r} &  705$^2$ &80.5 & - \\
  C-FRCNN\cite{chen2018context} & 705$^2$ & 82.2 & - \\
  Deformable ConvNet \cite{dai2017deformable} & 705$^2$ & 82.3 & 67.8\\
  CoupleNet\cite{zhu2017couplenet} & 705$^2$ & 82.7 & - \\
  FSN\cite{zhai2018feature} & 705$^2$ & 82.9 & - \\
  Deformable ConvNet v2 \cite{zhu2018deformable} & 705$^2$ & 84.9 & 73.5\\
  \midrule
 SNIPER  & 1915$^2$ & \textbf{86.6} & \textbf{80.5} \\ 
 \midrule
AutoFocus*& 860$^2$ & 85.8 & 79.5 \\ 
 \midrule
\multirow{2}{*}{AutoFocus} & 700$^2$ & 85.3 & 78.1 \\
 & 1250$^2$ & 86.5 & 80.2 \\
    
 \bottomrule
 \end{tabular}

\end{table}

\subsection{Comparison with other methods}\label{sssec:exp_final_comparison}
In this section, we compare our methods with other object detectors on COCO and Pascal VOC datasets. Table \ref{tab:final_coco} shows the results on COCO test-dev. One has to keep in mind that it is difficult to fairly compare different detectors as they differ in backbone architectures (like ResNet \cite{he2016deep}, ResNext \cite{xie2017aggregated}, Xception \cite{chollet2016xception}), pre-training data (\textit{e.g.} ImageNet-5k, JFT \cite{hinton2015distilling}, OpenImages \cite{openimages}), different structures in the underlying network (\textit{e.g} multi-scale features \cite{lin2017feature,najibi2017ssh}, deformable convolutions \cite{dai2017deformable}, heavier heads \cite{peng2017megdet}, anchor sizes, path aggregation \cite{liu2018path}), test time augmentations like flipping, mask tightening, iterative bounding box regression etc. In our comparison, we use a ResNet-101 backbone for all methods. Besides AP at different thresholds and different object sizes, we also show the average number of pixels processed by each method during inference.

First, among our methods, SNIPER outperforms its baseline SNIP, while improving the training speed noticeably. AutoFocus, on the other hand, exactly matches its baseline SNIPER, in terms of AP, however, reduces the number of pixels processed by more than 2X. In terms of absolute clock time, SNIPER processes 2.5 images per second. AutoFocus increases the speed to 6.4 images per second. To compare, RetinaNet with a ResNet-101 backbone and a FPN architecture processes 6.3 images per second on a P100 GPU (which is like Titan X), but obtains 37.8\% mAP \footnote{\url{https://github.com/facebookresearch/Detectron/blob/master/MODEL_ZOO.md}}. As can be seen, AutoFocus can effectively reduce the number of processed pixels further to $860^2$ while still achieving higher APs compared to multi-scale methods such as RepPoins+ \cite{yang2019reppoints} which on average processes around 11$\times$ more pixels.

 We also report results on the PASCAL VOC dataset in Table \ref{tab:final_voc}. To show the robustness of AutoFocus to its hyper-parameters, we use exactly the same hyper-parameters tuned for COCO (shown as AutoFocus*). While processing the same area as DeformableV2 \cite{zhu2018deformable}, AutoFocus achieves 4.6\% better AP at 0.7 IoU. It also matches the performance of SNIPER while being considerably more efficient.


\section{Conclusion}
We provided critical insights into the popular single-scale object-detection paradigm and highlighted some of its detrimental limitations. Carefully designed experiments showed that large scale-variation in object sizes adversely affects both the training and inference performance for object detection. Based on the characteristics of the human foveal-vision system and scale-space theory, scale-normalized image pyramids are proposed as an effective tool to tackle the aforementioned scale-variation and its effectiveness is showed on multiple popular object-detection systems. Generalizable guidelines are also provided to implement scale-normalization based on the input image, network architecture and objects of interest that can be further used for other applications as well. Our proposed technique to perform efficient spatial and scale-space sub-sampling of salient regions resulted in 3$\times$ faster training and 10$\times$ reduction in memory complexity which countered the increased computational complexity introduced by the scale-normalized image pyramid. The reduced memory complexity also enabled the use of batch-normalization which improved the results further, leading to state-of-the-art performance on the COCO benchmark. Finally, we presented an active foveal vision-system that processes the image pyramid in a coarse-to-fine manner to predict the location of object-like regions in the finer resolution scales, which speeds up inference by 3$\times$ resulting in \emph{near} real-time detection on commodity GPUs.

\FloatBarrier

\ifCLASSOPTIONcompsoc
  \section*{Acknowledgments}
  The authors would like to thank an Amazon Machine Learning gift for the AWS credits used for this research. The research is based upon work supported by the Office of the Director of National Intelligence (ODNI), Intelligence Advanced Research Projects Activity (IARPA), via DOI/IBC Contract Numbers D17PC00287 and D17PC00345. The U.S. Government is authorized to reproduce and distribute reprints for Governmental purposes not withstanding any copyright annotation thereon. Disclaimer: The views and conclusions contained herein are those of the authors and should not be interpreted as necessarily representing the official policies or endorsements, either expressed or implied of IARPA, DOI/IBC or the U.S. Government.
\else
  \section*{Acknowledgment}
\fi

\ifCLASSOPTIONcaptionsoff
  \newpage
\fi

\bibliographystyle{IEEEtran}
\bibliography{IEEEabrv,bare_jrnl_compsoc}

\begin{IEEEbiography}[{\includegraphics[width=1in,height=1.25in]{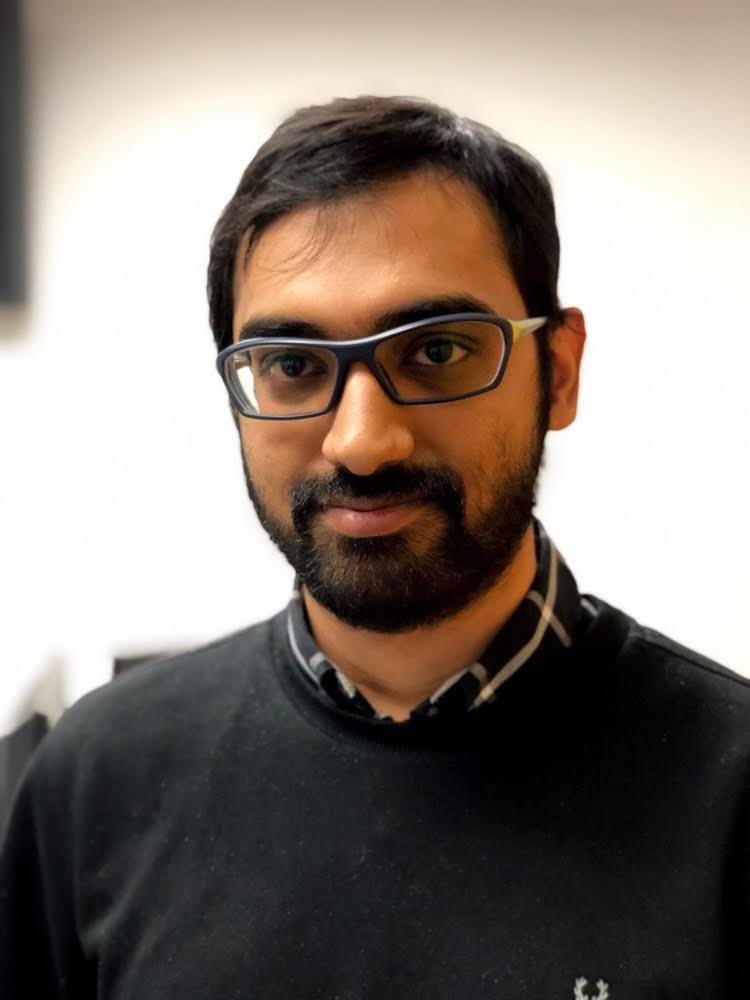}}]{Bharat Singh}
received his Bachelor's and Master's degree in computer science from Indian Institute of Technology Madras in 2013, and the Ph.D. degree in computer science from University of Maryland College Park in 2018. His research interests include computer vision and machine learning.

\end{IEEEbiography}
\begin{IEEEbiography}[{\includegraphics[width=1in,height=1.25in]{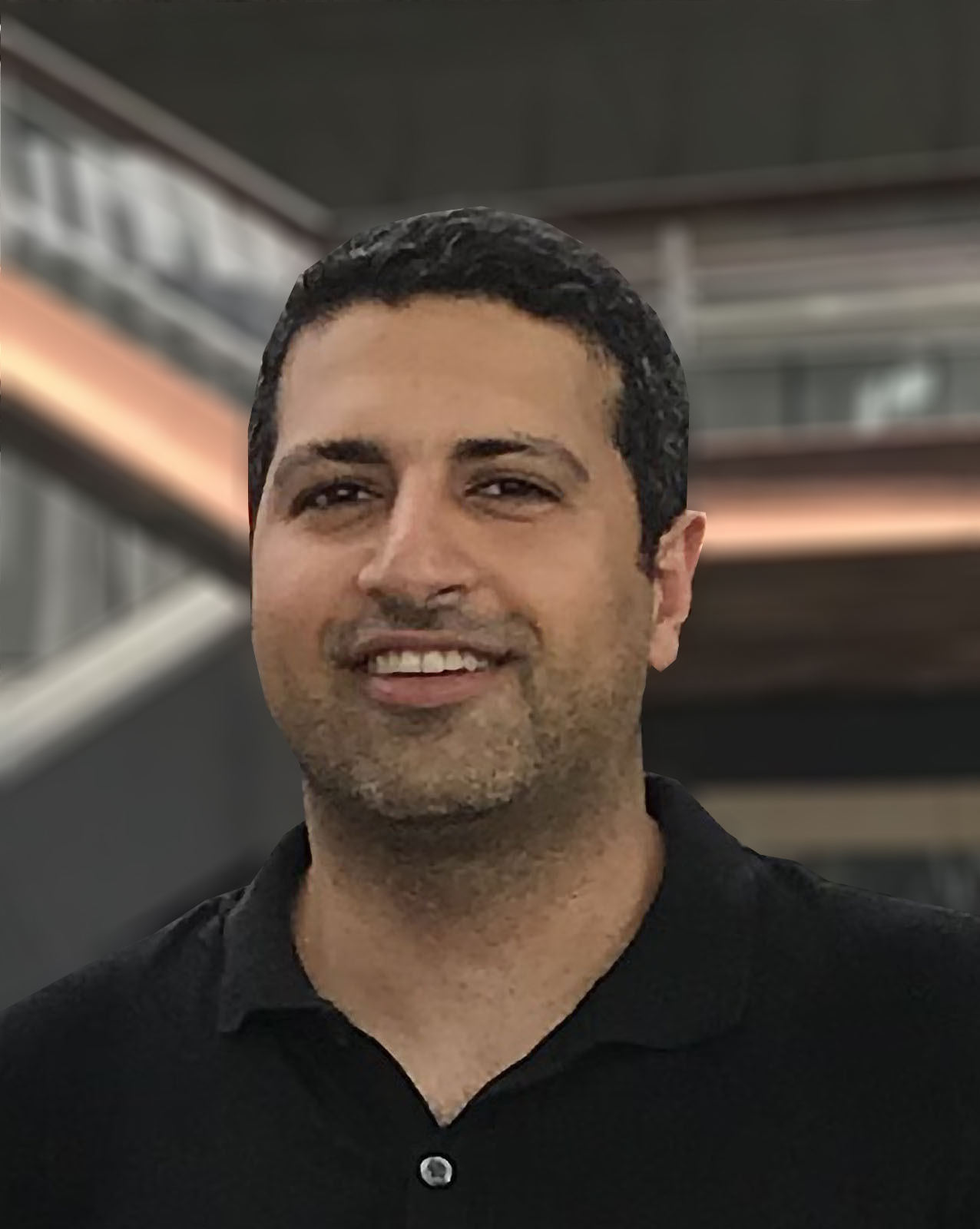}}]{Mahyar Najibi}
received his B.Sc. in Computer Engineering from Iran University of Science and Technology, his M.Sc. in Artificial Intelligence from the Sharif University of Technology, and his Ph.D. in Computer Science from University of Maryland, College Park in 2020. His research interests lie in the intersection of computer vision and machine learning.

\end{IEEEbiography}
\begin{IEEEbiography}[{\includegraphics[width=1in,height=1.25in]{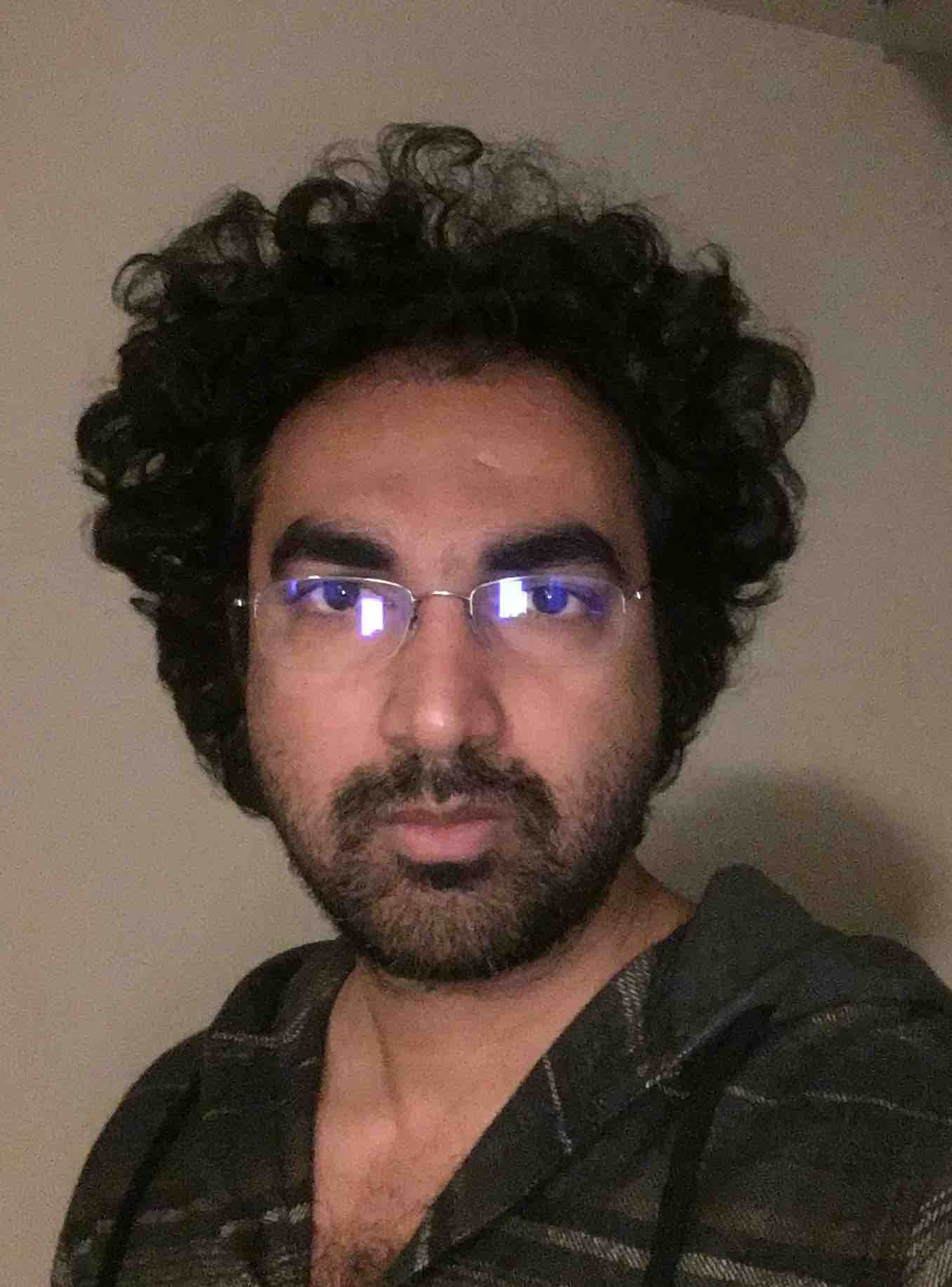}}]{Abhishek Sharma}
received his BS degree in electrical engineering from Indian Institute of Technology Roorkee in 2010, and the Ph.D. degree in computer science from University of Maryland College Park in 2015. His research interest includes visual automation, biometrics and machine learning.
\end{IEEEbiography}
\begin{IEEEbiography}[{\includegraphics[width=1in,height=1.25in,clip,keepaspectratio]{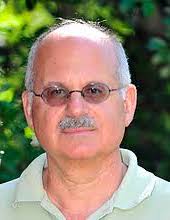}}]{Larry S. Davis} received his B.A. from Colgate University in 1970 and his M.S. and Ph.D. in Computer Science from the University of Maryland in 1974 and 1976 respectively. From 1977-1981 he was an Assistant Professor in the Department of Computer Science at the University of Texas, Austin. He returned to the University of Maryland as an Associate Professor in 1981. From 1985-1994 he was the Director of the University of Maryland Institute for Advanced Computer Studies. He was Chair of the Department
of Computer Science from 1999-2012. He is currently a Professor in the Institute and the Computer Science Department. He was named a Fellow of the IEEE in 1997 and of the ACM in 2013.
\end{IEEEbiography}
\end{document}